\documentclass{article}

\usepackage[numbers]{natbib}
    \usepackage[preprint]{neurips_2025}

\usepackage{booktabs}  
\usepackage{amsfonts}  
\usepackage{pifont}    
\usepackage{graphicx}  
\usepackage[utf8]{inputenc} 
\usepackage{pifont}           
\newcommand{\cmark}{\ding{51}}
\newcommand{\xmark}{\ding{55}}

\usepackage[T1]{fontenc}    
\usepackage{hyperref}       
\usepackage{url}            
\usepackage{booktabs}       
\usepackage{amsfonts}       
\usepackage{nicefrac}       
\usepackage{microtype}      
\usepackage{booktabs}  
\usepackage{makecell}  
\usepackage{subcaption}
\usepackage{graphicx}
\usepackage[ruled,vlined]{algorithm2e}
\usepackage{amsmath}

\usepackage{algorithmic}

\usepackage{amsmath}
\usepackage{amsthm}

\usepackage{wrapfig}
\usepackage{multirow}

\usepackage{booktabs}
\usepackage[table,xcdraw]{xcolor}  
\usepackage{graphicx}

\definecolor{fixedgray}{gray}{0.85}
\definecolor{varyinggreen}{RGB}{199,233,192}
\definecolor{highlightgreen}{RGB}{120,200,120}
\definecolor{lightblue}{RGB}{180,210,255}
\definecolor{lightred}{RGB}{255, 205, 210}     
\definecolor{lightorange}{RGB}{255, 224, 178}  

\definecolor{lightpurple}{RGB}{220, 200, 255}
\definecolor{codegray}{rgb}{0.5,0.5,0.5}
\definecolor{codeblue}{rgb}{0.2,0.4,0.8}
\definecolor{backcolor}{rgb}{0.95,0.95,0.92}

\definecolor{CustomOrange}{RGB}{242, 166, 100}
\definecolor{CustomBlue}{RGB}{60, 100, 220}
\definecolor{CustomLightPurple}{RGB}{205, 160, 230}

\usepackage[utf8]{inputenc} 
\usepackage[T1]{fontenc}    
\usepackage{hyperref}       
\usepackage{url}            
\usepackage{booktabs}       
\usepackage{amsfonts}       
\usepackage{nicefrac}       
\usepackage{microtype}      
\usepackage{listings}
\usepackage{subcaption}
\usepackage{graphicx}

\lstdefinestyle{pythonstyle}{
    backgroundcolor=\color{backcolor},
    commentstyle=\color{black}\itshape,
    numberstyle=\tiny\color{gray},
    basicstyle=\ttfamily\scriptsize,
    breaklines=true,             
    breakatwhitespace=true,      
    breakautoindent=false,       
    breakindent=0pt,             
    columns=flexible,            
    keepspaces=true,             
    frame=single,
    numbers=none,
    numbersep=5pt,
    language=Python,
    showstringspaces=false,
    tabsize=4
}
\lstdefinestyle{pythoncolorstyle}{
    backgroundcolor=\color{backcolor},
    commentstyle=\color{black}\itshape,
    keywordstyle=\color{codeblue}\bfseries,
    numberstyle=\tiny\color{gray},
    stringstyle=\color{red},
    basicstyle=\ttfamily\scriptsize,
    breaklines=true,             
    breakatwhitespace=true,      
    breakautoindent=false,       
    breakindent=0pt,             
    columns=flexible,            
    keepspaces=true,             
    frame=single,
    numbers=none,
    numbersep=5pt,
    language=Python,
    showstringspaces=false,
    tabsize=4
}

\title{RoboMoRe: LLM-based Robot Co-design via Joint Optimization of Morphology and Reward}

\author{
Jiawei Fang \\
Department of Mechanical Engineering \\
University of California, Berkeley \\
\texttt{jiaweif@berkeley.edu} \\
\And
Yuxuan Sun \\
Department of Mechanical Engineering \\
University of California, Berkeley \\
\texttt{yuxuan\_sun@berkeley.edu} \\
\And
Chengtian Ma \\
Department of Mechanical Engineering \\
University of California, Berkeley \\
\texttt{chtianma@gmail.com} \\
\And
Qiuyu Lu\textsuperscript{*} \\
Department of Mechanical Engineering \\
University of California, Berkeley \\
\texttt{qiuyulu@berkeley.edu} \\
\And
Lining Yao\thanks{corresponding authors} \\
Department of Mechanical Engineering \\
University of California, Berkeley \\
\texttt{liningy@berkeley.edu} \\
}

\begin{document}

\maketitle
\begin{abstract}

Robot co-design, jointly optimizing morphology and control policy, remains a longstanding challenge in the robotics community, in which many promising robots were generated. However, a key limitation lies in its tendency to converge to sub-optimal designs due to the use of fixed reward functions, which fail to explore the diverse motion modes suitable for different morphologies. Here we propose RoboMoRe, a large language model (LLM)-driven framework that integrates morphology and reward shaping for co-optimization within the robot co-design loop. RoboMoRe performs a dual-stage optimization: in the coarse optimization stage, an LLM-based diversity reflection mechanism is integrated to generate both diverse and high-quality morphology–reward pairs and efficiently explore their distribution. In the fine optimization stage, top candidates are iteratively refined through alternating LLM‑guided reward and morphology gradient updates. RoboMoRe could optimize both efficient robot morphology and corresponding suited motion behaviors by reward shaping. The results demonstrate that without any task-specific prompting or predefined reward and morphology templates, RoboMoRe significantly outperform human-engineered design results and competing methods in eight different tasks. 

\end{abstract}

\section{Introduction}
\label{sec:intro}

\begin{figure*}[htbp]
  \centering
  \includegraphics[width=1.0\textwidth]{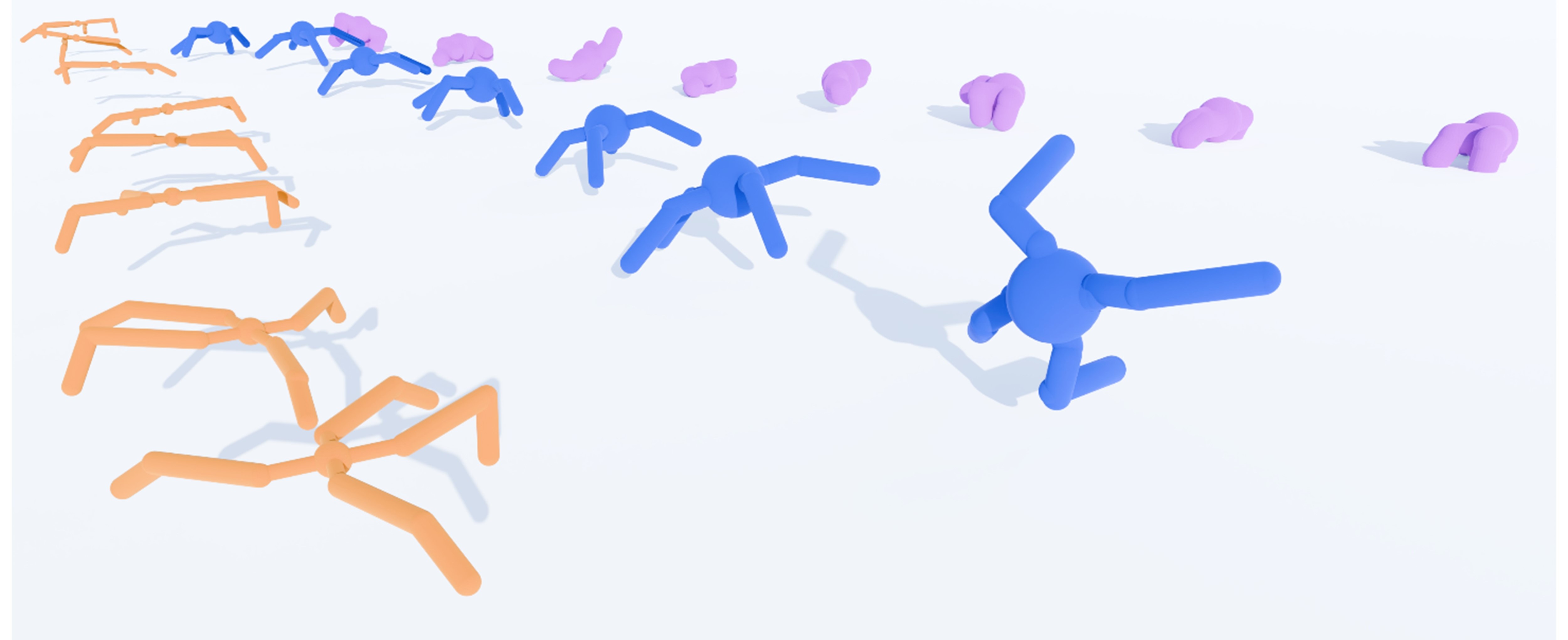} 
  \caption{ \textbf{Diverse motion behaviors and robot morphologies generated by RoboMoRe}. \textcolor{CustomOrange}{\textbf{Orange}}: A bio-inspired long-legged robot crawling forward in a climber posture.  
\textcolor{CustomBlue}{\textbf{Blue}}: A uniform robot advancing by jumping. 
\textcolor{CustomLightPurple}{\textbf{Purple}}: A low-center-of-gravity robot rolling forward along the ground.
}
  \label{fig:Fig_teaser} 
 
\end{figure*}

One of the main goals of artificial intelligence is to develop effective approaches for the creation of embodied intelligent systems \cite{steels2018artificial}. Inspired by natural organisms, where body structure and brain are two key factors for completing any task in a real environment, a successful intelligent robot typically requires concurrently optimizing its structure design and control mechanism \cite{saridis1983intelligent}. Such a co-design problem has been a long-standing key challenge in the robotics and machine learning communities \cite{wang2023softzoo, ma2021diffaqua, xu2021multi}.  

Although many existing approaches have achieved promising results, they typically rely on a fixed reward function, which significantly limits the potential of robot co-design \cite{lu2025bodygen, zhao2020robogrammar}. Under such constraints, robots are only capable of learning a narrow range of motion behaviors tied to a single, static objective. \textcolor{black}{For robots with varying morphologies, a single, uniform reward function severely constrains improvements in locomotion performance and the exploration of diverse motion modalities\cite{cui2025grove,wang2024towards,han2024generating}.}

To address this limitation, we introduce \textbf{RoboMoRe}—a novel LLM-based robot co-design framework that jointly optimizes both \textbf{Robo}t \textbf{Mo}rphology and \textbf{Re}ward functions. The core philosophy of RoboMoRe \textcolor{black}{is to identify optimal reward functions tailored to each robot morphology.}  By tailoring reward functions to match specific morphologies, we can unlock a wider variety of motion behaviors and boost overall performance. For example, a long-legged robot might benefit from a reward that encourages jumping, while a low-profile robot might perform better with a reward that promotes rolling. Please refer to Appendix \ref{supp:motivation} for more details.

Our framework follows a Coarse-to-Fine Optimization paradigm. In the Coarse Optimization phase, we introduce a simple and effective Diversity Reflection mechanism in which the LLM observes previously generated morphology-reward pairs and proposes new, diverse candidates for subsequent grid search. This promotes the exploration of the joint distribution space and ensures a wide coverage of viable design alternatives. Then, in the fine optimization phase, we iteratively refine the most promising candidates in an alternating optimization scheme \cite{bezdek2003convergence}. Specifically, the LLM first refines the reward function based on existing samples, then adjusts the morphology accordingly. This reward-morphology alternating cycle continues until convergence on high-performing design pairs. By fully harnessing the power of LLMs, we aim to inspire a new direction in robot co-design that tightly couples morphology and reward shaping.

In summary, our contributions are three-fold:
\begin{enumerate}
    \item We propose RoboMoRe, a novel co-design framework that leverages LLM to jointly optimize robot morphology and reward functions.
    \item We adopt a Coarse-to-Fine optimization paradigm that combines Diversity Reflection for grid search and alternating refinement for precise optimization.
    \item Our quantitative and qualitative evaluations show that RoboMoRe significantly outperforms both human-designed and competing automated methods across eight representative design tasks. 
\end{enumerate}

\section{Related Work}
\label{sec:related}
\paragraph{Robot Co-design}
Early evolutionary robotics demonstrated that co-evolving morphology and control policies outperform sequential design, highlighting the interdependence of “body” and “brain” in agents \cite{bravo2020one}. Early work framed robot co-design as a graph search problem and addressed using the advanced evolutionary algorithms, yielding progressively more intriguing result \cite{sims2023evolving, ha2019reinforcement, pathak2019learning, wang2019neural, zhao2020robogrammar}. However, as both kinematic complexity and task difficulty grow, approaches such as evolutionary algorithms, Bayesian optimization, and deep reinforcement learning must contend with vast design spaces that demand substantial computational resources and more efficient search strategies, thereby constraining their development. Recently, large language models (LLMs) hold significant promise for transforming the robot design process thanks to their strong in-context learning capabilities and extensive prior knowledge. However, research on LLM-based robot co-design remains limited to only a few early exploratory studies  \cite{zhang2024cuda, qiu2024robomorph,lehman2023evolution,songlaser}. Among these, LASER \cite{songlaser} has shown promising results in designing soft voxel robots (SVR), particularly with regard to reflecting diversity. However, it still relies on carefully crafted task-specific prompts and a highly narrow method for diversity reflection which can only be applied to SVR,  greatly limits its generalizability to other types of robots. In contrast, our work introduces highly general prompts and a more flexible approach to diversity reflection. Most importantly, existing methods share a critical limitation with fixed reward strategies, which significantly hinders their effectiveness in robot co-design.

In general, while these methods demonstrate potential, previous approaches have not incorporated reward shaping. By fixing the reward function and only optimizing robot morphology—i.e., \textit{Fixed Reward, Varying Morphology}—they have restricted the possibilities of true co-design. In our paper, we argue that reward shaping can encourage different robots to develop distinct and diverse movement strategies, ultimately resulting in superior performance.

\paragraph{Reward Shaping}
Reward shaping is a pivotal component in the development of robotic control strategies. Early reward engineering methods predominantly relied on  manual trial-and-error tuning \cite{knox2023reward} and inverse reinforcement learning (IRL) \cite{ho2016generative}. Manual tuning is time-consuming and requires substantial domain expertise, whereas IRL demands costly expert demonstrations and frequently produces opaque, black-box reward functions, thereby limiting its practical applicability.  Subsequent researches  have investigated automated reward optimization via evolutionary algorithms \cite{faust2019evolving,chiang2019learning}, but these efforts were typically limited to task-specific implementations that only tuned parameters within predefined reward templates. More recently, large language models (LLMs) have been used to generate reward functions for new tasks. For instance, Eureka \cite{ma2023eureka} provides a general framework that, without any task-specific prompting or pre-defined reward templates, can generate reward functions that outperform human-engineered ones. Similarly, Text2Reward \cite{yang2024text2reward} shows that LLM-based reward shaping can produce symbolic, interpretable rewards and enable robots to learn novel locomotion behaviors (e.g., \textit{Hopper backflip, Ant lying down}).

Inspired by these advances, we introduce LLM-driven reward shaping into robot co-design. By generating diverse reward functions with LLMs, RoboMoRe enables robots with varying morphologies to discover novel motion strategies, significantly enhancing their performance.

\vspace{-5pt}
\begin{table}[ht]
\centering
\caption{\textbf{Comparison of RoboMoRe with related literature.}}
\label{tab:literature_comparison}
\resizebox{\textwidth}{!}{
\begin{tabular}{c|cccccc}
\toprule
\textbf{Literature} & \textbf{Method} & \textbf{Mor.\ Design} & \textbf{Reward Shaping} & \textbf{Diversity} & \textbf{Iterations} & \textbf{Optimum} \\
\midrule

EvoGym~\cite{bhatia2021evolution} & BO, EA             & \cmark & \xmark & N/A & Slow & Local \\
RoboGrammar~\cite{zhao2020robogrammar} & Graph Search  &\cmark & \xmark & N/A & Slow & Local \\
RoboMorph~\cite{qiu2024robomorph} & LLM            & \cmark & \xmark & Off & Fast & Local \\
LASER~\cite{songlaser}          & LLM            & \cmark & \xmark & On  & Fast & Local \\
Eureka~\cite{ma2023eureka}      & LLM            & \xmark & \cmark & Off & Fast & Local \\
\textbf{RoboMoRe (Ours)}        & LLM            & \cmark & \cmark & On  & Fast & Near Global \\
\bottomrule
\end{tabular}%
}
\end{table}

\section{Problem Definition (Robot Co-design)}
The Robot Co-design problem involves jointly optimizing a robot’s morphology and control policy to maximize its performance in a given environment. It is defined as a tuple $
P =\langle \mathcal{M}, \Theta, \mathcal{R}, \mathcal{A}, F \rangle
$, where $\mathcal{M}$ is the environment model, $\Theta$ is the robot morphology design space, and $\mathcal{R}$ is the space of reward functions. $\mathcal{A}(\theta, R): \Theta \times \mathcal{R} \to \Pi$  is the co-design algorithm that takes a robot design $\theta$ and reward function $R$, then learns an optimal policy $\pi$. $F$ is the fitness function which measures real-world performance but can only be accessed through policy execution. 

Existing works aim to solve this problem via searching the optimal robot design $\theta^*$ that lead to the highest performance given a fixed reward function $R_0$:
\begin{equation}
    \theta^* = \arg\max_{\theta \in \Theta} F( \pi_{\theta, R_0}),
\end{equation}
However, this can lead to local optimum due to limitation of $R_0$. To address this, our objective is to jointly optimize the robot morphology $\theta^*$ and reward function $R^*$ that lead to the global optimal performance:
\begin{equation}
    \theta^*, R^* = \arg\max_{\theta \in \Theta, R \in \mathcal{R}} F(\pi_{\theta, R}),
\end{equation}
where the optimized policy is obtained by:

\begin{equation}
    \pi_{\theta, R} := \mathcal A(\theta, R).
\end{equation}

\section{Method}

\begin{figure*}[htbp]
  \centering
  \includegraphics[width=1.0\textwidth]{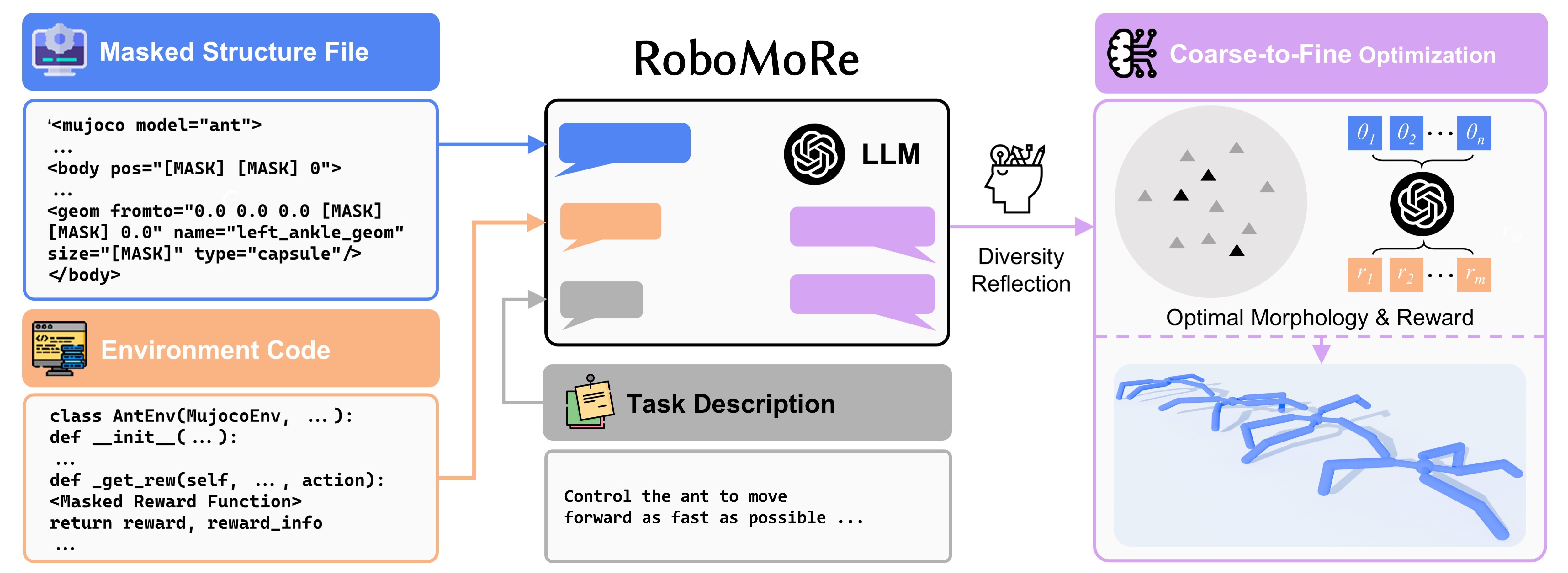} 
  \caption{\textbf{The overall pipeline of RoboMoRe}. The LLM is provided with the task description, environment code, and masked structure file as the input prompt. The environment code specifies the simulation variables to write reward functions, while the masked structure file provides masked morphology parameters of robots for the LLM to complete. Our algorithm follows a coarse-to-fine strategy. First, the LLM leverages diversity reflection to generate a diverse set of reward functions and morphologies, enabling broad exploration of the reward-morphology design space (the coarse optimization phase). Then, through alternating optimization between morphology and reward, the framework iteratively refines both to arrive at an optimal morphology-reward pair (the fine optimization phase).}
  \label{fig:Fig_framework} 

\end{figure*}

In this section, we formally introduce the RoboMoRe pipeline. \textcolor{black}{See Fig. \ref{fig:Fig_framework} for an overview.} Traditional robot co-design approaches typically optimize morphology under a fixed reward function, which constrains the search space and often results in sub-optimal behaviors misaligned with the robot’s morphology \cite{lu2025bodygen}. 



To address this limitation, RoboMoRe introduces a co-optimization framework that simultaneously explores morphology and reward function space, enabling both components to co-evolve toward globally optimal performance. By dynamically adapting reward functions to \textcolor{black}{suit varying} morphologies, RoboMoRe unlocks diverse and morphology-aware motion behaviors, maximizing overall efficiency. For instance, a flexible robot equipped with elongated limbs should be encouraged to optimize its movement through jumping-based locomotion, leveraging its mechanical advantage. In contrast, a heavier robot with short limbs may be better suited for rolling-based movement, as this motion minimizes energy expenditure while maximizing travel distance. This adaptive approach allows RoboMoRe to systematically align each morphology with its most compatible motion strategy, resulting in superior performance across a wide range of robotic configurations.

\subsection{Coarse Optimization: Grid Search with Diversity Reflection}
The goal of coarse optimization is to approximate the distribution of morphology-reward pairs near the global optimum. To achieve this, the generated samples must satisfy two key criteria: 1) \textit{diversity}, meaning the samples should be sufficiently varied to avoid premature convergence to local optima; and 2) \textit{high-quality}, indicating that the samples should also cluster near the global optimum to improve sampling efficiency.

Traditional approaches typically rely on random sample (e.g., sampling initial candidates from gaussian distribution) \cite{ha2019reinforcement, bhatia2021evolution}, which although ensures diversity, often fails to generate samples sufficiently close to the global optimum. As a result, these methods require hundreds of thousands of optimization steps to converge, making them computationally expensive and inefficient. To address this, RoboMoRe leverage LLM and a simple Diversity Reflection mechanism to generate high-quality and diverse samples. We leverage the advanced reasoning, comprehension, and design capabilities of LLMs to generate high-quality samples. For morphology design, we input the task description alongside a masked robot structure file (e.g., an XML file for the MuJoCo simulation environment) into the LLM, prompting it to generate a novel set of robot design parameters. Crucially, we mask all morphology-related parameters (e.g., limb lengths) within the structure file to eliminate bias stemming from human-designed priors, ensuring a fair and unbiased evaluation of the generated designs. For reward functions, following the Eureka, we directly ingest the raw environment source code—excluding any pre-existing reward functions—as contextual information. This enables the LLM to synthesize candidate reward functions that are both tailored to the task and executable in Python, seamlessly integrating into the simulation environment. We impose a strict output format to enforce structural consistency and coherence. Although task-specific prompt engineering could enhance the quality of generated samples \cite{songlaser}, RoboMoRe adapts a highly generalizable prompt to ensure scalability in diverse tasks. Please refer to Appendix \ref{supp:prompts} for more details in our prompt.

\begin{figure*}[htbp]
  \centering
  \includegraphics[width=1.0\textwidth]{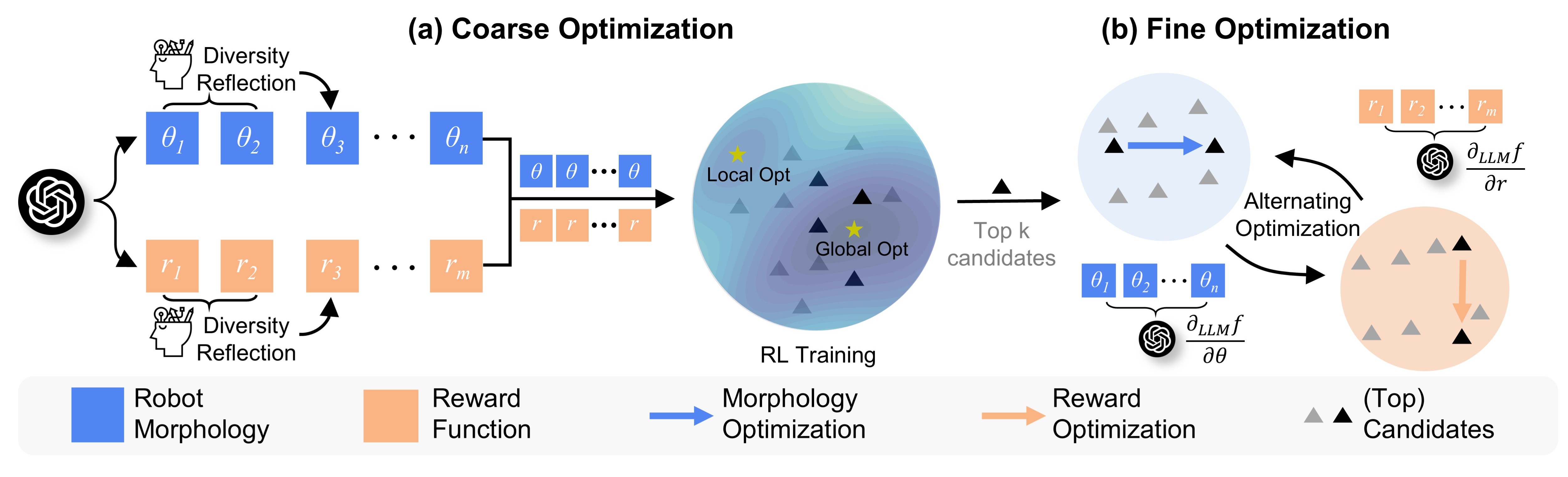} 
  \caption{\textbf{Overview of Coarse-to-Fine algorithm.} (a) In the coarse optimization stage, the LLM first generates diverse reward functions and robot morphologies using a Diversity Reflection mechanism. These reward-morphology pairs are then trained via reinforcement learning to explore and identify the most suitable motion behavior for each robot. (b) In the fine optimization stage, the LLM employs Alternating Optimization to jointly refine both the robot's reward and morphology, leading to an optimized design and control strategy.}
  \label{fig:Fig_algorithm} 
\end{figure*}

Although LLM can generate high-quality samples, our reward-morphology optimization theory suggests that excessive repetition among samples can lead to sub-optimal local maximum. To address this issue, we propose a simple and effective mechanism called Diversity Reflection to significantly enhance the diversity of LLM-generated designs. The core idea of diversity reflection is intuitive, as illustrated in Fig.~\ref{fig:Fig_algorithm}. For each newly proposed morphology or reward function, the LLM reflects on previously generated samples and deliberately produces a new design that maximizes diversity relative to past candidates. Owing to its simplicity, Diversity Reflection is task-agnostic and can potentially be seamlessly extended to other domains without any modification. Experimental results in Sec. \ref{sec:diversity} demonstrate that diversity reflection consistently improves both sample diversity and optimization performance across diverse tasks.

\subsection{Fine Optimization: Morphology Reward Alternating Optimization}

After obtaining an approximate distribution of morphology and reward function space, we employ an alternating optimization strategy to iteratively refine morphology and reward function, progressively converging toward the global optimum. Specifically, we filter the best-performing candidates from the previous coarse optimization round and feed them into the LLM. The LLM first analyzes the robot morphology samples, identifying morphology optimization momentum $\frac{\partial_{LLM}f}{\partial\theta}$ to iteratively enhance morphology. Subsequently, LLM will examine the reward function samples, refining the reward design to better align with the evolving morphology according to reward optimization momentum$\frac{\partial_{LLM}f}{\partial r}$. Additionally, we retain and explore the most promising coarse optimization results, expanding the search space and uncovering further optimization possibilities, thereby facilitating a more comprehensive search for the global optimum. Please see Appendix \ref{supp:algorithm} for the algorithm details.

\vspace{-0.5cm}
\section{Experiment Setup}
\paragraph{Environments} 
We implemented our environments using the MuJoCo Gym simulator, covering eight distinct robot design tasks: Ant, Ant-Powered, Ant-Desert, Ant-Jump, Hopper, Half-Cheetah, Swimmer and Walker . These environments encompass diverse physical world types (2D, 3D), environment tasks, search space complexities, ground terrains, and gear powers. For morphology design, we mask certain parameters in the MuJoCo XML files to prevent the LLM from gaining prior knowledge of the robot's design. For reward function design, the LLM is provided with the official environment code with the reward functions removed. Regarding task descriptions, we use the official descriptions from the environment repository whenever available. Detailed information on robot design morphologies, reward functions, and task descriptions can be found in Appendix \ref{supp:env}.
\paragraph{Baselines} 
We consider 4 strong baselines to highlight performance of RoboMoRe: 1) \textbf{Bayesian Optimization (BO)} is a classic algorithm designed to optimize expensive-to evaluate functions \cite{snoek2012practicalbayesianoptimizationmachine,rasmussen2003gaussian}. It employs a probabilistic model (e.g. Gaussian Process) as a surrogate for the objective function and determines where to sample based on predicted mean and uncertainty. 2) \textbf{Eureka} is a general LLM-based reward shaping framework \cite{ma2023eureka}. Similar to an evolutionary algorithm, It uses an LLM to generate a large number of samples, select the best-performing ones, and acts as a heuristic operator for mutation. 3) \textbf{Eureka (Mor.)} is a modified version of the original Eureka framework, specifically focused on robot morphology design. Unlike the standard reward shaping Eureka, Eureka (Mor.) leverages the LLM’s capability to explore and optimize robot morphology parameters. 4) \textbf{Human} represents the template morphology and reward functions from official Mujoco website \cite{towers2024gymnasium}. As these morphologies and reward functions are written by active reinforcement learning researchers who designed the tasks, these reward functions represent the outcomes of expert-level human reward engineering.


\paragraph{Evaluation Metrics}
In all experiments, we use a normalized metric \textbf{efficiency} to evaluate performance, which is defined as division of fitness and robot volume. 
This choice is crucial for stimulating the LLM’s material-aware design capabilities rather than brute-force scaling. We compute volume using custom scripts for each type of robot. While fitness is also reported (e.g., distance for locomotion, height for jumping tasks), \textbf{efficiency} serves as the core metric in our optimization iterations. Please refer to Appendix \ref{supp:Implementation Details} for more implementation details.

\section{Results}
\subsection{Comparison with Competing Methods}

\begin{figure*}[htbp]
  \centering
  \includegraphics[width=1.0\textwidth]{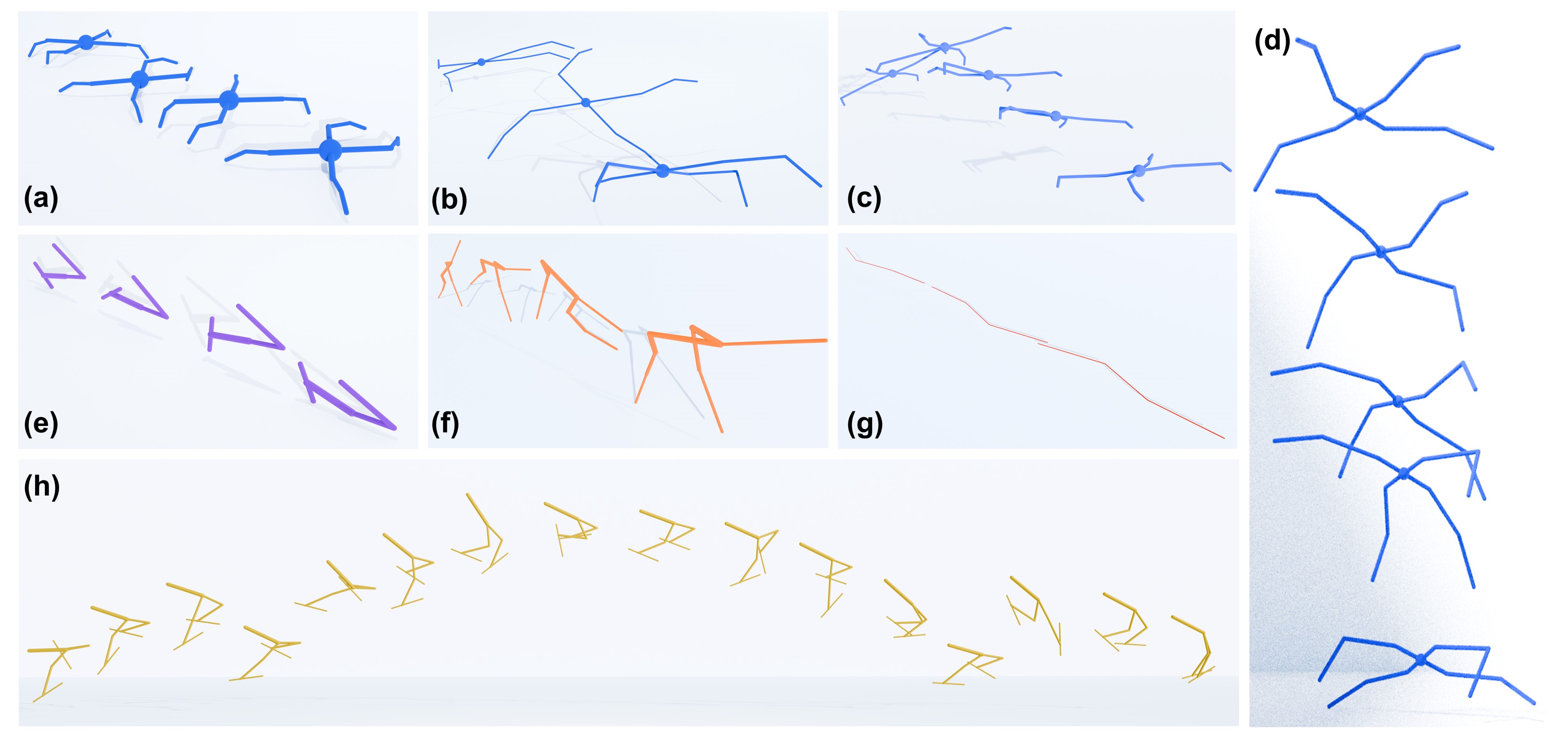} 
  \caption{\textbf{Optimal designs generated by RoboMoRe.} (a) Ant. (b) Ant-Powered, where motor power is doubled. (c) Ant-Desert, with the terrain replaced by desert. (d) Ant-Jump, where the task is changed from locomotion to jumping. (e) Hopper. (f) Half-Cheetah. (g) Swimmer. (h) Walker.}
  \label{fig:Fig_4} 
  \vspace{-.3cm}
\end{figure*}

\paragraph{RoboMoRe outperforms human design robots and competing methods.} Table \ref{table:comparison} quantitatively compares the results of our method and other strong baselines. The corresponding morphology and reward function can be found in Appendix \ref{supp:Comparison of optimal morphology design via different methods}. It is evident that RoboMoRe consistently generates highly performant designs, significantly outperforming both human-generated and baseline method designs. Notably, RoboMoRe demonstrates a substantial advantage in efficiency, which can be attributed to its ability to generate more structurally efficient robotic designs (Fig. \ref{fig:Fig_4}). We observe a remarkable performance gap between human-designed results and our method, especially in terms of efficiency. We believe this stems from the fact that humans are less likely to approach near-optimal designs for complex tasks, especially for efficiency, leaving substantial room for improvement.  In addition, our competitors underperform mainly for two reasons. First, LLM framework Eureka and Eureka (Mor.) fail to consider the importance of diversity during optimization, often producing repetitive and suboptimal offsprings. We further analyze the importance of diversity in Section~\ref{sec:diversity}. 
 Second, both methods focus solely on optimizing either the reward function or morphology, which limits their potential and results in suboptimal performance. We further analyze the importance of both reward shaping and morphology design in Section~\ref{sec:reward_shaping}. In addition, we conducted robustness analyses on two cases: Ant-Powered and Ant-Desert. For Ant-Powered, we varied the gear power configurations, while for Ant-Desert, we evaluated performance across different terrains. Detailed results are provided in Appendix \ref{supp:Robustness across Different Gear Powers} and Appendix \ref {supp:Robustness across Different Terrains}.

\begin{table}[ht]
\centering
\caption{\textbf{Performance comparison of human and competing methods across eight tasks.} We report both efficiency and fitness metrics. Note that we set efficiency as the optimization goal instead of fitness in implementation.}
\label{table:comparison}
\resizebox{\textwidth}{!}{%
\begin{tabular}{llrrrrrrrr}
\toprule
\textbf{Metric} & \textbf{Method} & \textbf{Ant} & \textbf{Ant-Powered} & \textbf{Ant-Desert} & \textbf{Ant-Jump} & \textbf{Hopper} & \textbf{Half-Cheetah} & \textbf{Swimmer} & \textbf{Walker} \\
\midrule
\multirow{5}{*}{\makecell[l]{\textbf{Efficiency} \(\uparrow\) \\ \textbf{(Fitness/Volume)}}}
    & Bayesian Optimization & 707.13 & 74.52 & 223.61 & 3.53 & 128.00 & 4242.89 & 57.53 & 111.95 \\
    & Eureka & 121.54 & 24.91 & 49.82 & 7.60 & 428.11 & 12,157.92 & 23.96 & 139.73 \\
    & Eureka (Mor.) & 5,203.82 & 1,160.94 & 2.51 & 229.55 & 609.04 & 15,531.26 & 16,335.19 & 1,946.60 \\
    & Human & 68.22 & 26.13 & 29.84 & 6.83 & 433.69 & 11,975.34 & 18.86 & 170.29 \\
    & \textbf{RoboMoRe} & \textbf{31,038.41} & \textbf{32657.10} & \textbf{36,995.77} & \textbf{902.76} & \textbf{3,776.66} & \textbf{495,373.71} & \textbf{57,627.40} & \textbf{6,665.85} \\
\midrule
\multirow{5}{*}{\textbf{Fitness}\(\uparrow\)}
    & Bayesian Optimization & 390.14 & 41.11 & 100.49 & 1.95 & 2.75 & 194.82 & 8.66 & 2.54 \\
    & Eureka & 8.81 & 4.54 & 9.08 & 1.38 & 6.77 & 257.55 & 2.56 & 3.92 \\
    & Eureka (Mor.) & 143.07 & 41.52 & 0.08 & 5.33 & 1.61 & 145.09 & 89.04 & 2.91 \\
    & Human & 12.43 & 4.76 & 5.44 & 1.24 & 6.86 & 253.69 & 2.01 & 4.78 \\
    & \textbf{RoboMoRe} & \textbf{165.18} & \textbf{70.30} & \textbf{39.60} & \textbf{1.48} & \textbf{15.10} & \textbf{135.71} & \textbf{109.35} & \textbf{4.17} \\
\bottomrule
\end{tabular}
}
\end{table}

\subsection{Effectiveness of Diversity Reflection}
\label{sec:diversity}
Based on our main assumption, we formulate co-design as a joint optimization of morphology and reward. Therefore, to efficiently reach the global optimum, the initial solution set must satisfy two key properties: (1) \textit{high-quality}, denoting samples that demonstrate strong task performance and exhibit proximity to the global optimum in the design landscape; and (2) \textit{diversity}, denoting a sufficient variation among samples to mitigate premature convergence to local optima and facilitate broader exploration of the design space. Experimental results demonstrate that the diversity reflection mechanism empowers RoboMoRe to synthesize solutions that are both high-performing and structurally diverse, thereby enabling more comprehensive and effective exploration of the design landscape.

\begin{table*}[ht]
\centering
\caption{\textbf{Performance comparison of RoboMoRe variants in terms of efficiency, morphology diversity, and reward diversity.} For clarity, only results from the coarse optimization stage are reported.}
\label{tab:diversity_reflection_quantitative}
\resizebox{\textwidth}{!}{
\begin{tabular}{llrrrrrrrr}
\toprule
\textbf{Metric} & \textbf{Method} &\textbf{ Ant} & \textbf{Ant-Powered} & \textbf{Ant-Desert} & \textbf{Ant-Jump} & \textbf{Hopper} & \textbf{Half-Cheetah} & \textbf{Swimmer} & \textbf{Walker} \\
\midrule

\multirow{3}{*}{\textbf{Efficiency \(\uparrow\)}} 
& RoboMoRe (Random Sample) & 12.91 & 11.83 & 12.91 & 3.70 & 86.39 & 1,451.44 & 116.44 & 42.17 \\
& RoboMoRe (w/o Diversity Reflection) & 136.91 & 45.56 & 52.44 & 32.36 & 111.91 & 1,536.94 & 2,338.88 & 87.54 \\
& \textbf{RoboMoRe (w/ Diversity Reflection)} & \textbf{487.13} & \textbf{398.60} & \textbf{151.52} & \textbf{49.41} & \textbf{375.52} & \textbf{9,557.07} & \textbf{4,607.74} & \textbf{149.54} \\
\midrule

\multirow{3}{*}{\makecell[l]{\textbf{Morphology Diversity}\\
\textbf{(Coefficient of Variation) \(\uparrow\)}}}
& RoboMoRe (Random Sample) & 0.51 & 0.50 & 0.49 & 0.48 & \textbf{0.55} & 0.99 & 0.41 & \textbf{0.60} \\
& RoboMoRe (w/o Diversity Reflection) & 0.44 & 0.46 & 0.57 & 0.44 & 0.53 & 0.82 & 0.29 & 0.43 \\
& \textbf{RoboMoRe (w/ Diversity Reflection)} & \textbf{0.64} & \textbf{0.70} & \textbf{0.64} & \textbf{0.66} & 0.54 & \textbf{1.38} & \textbf{0.61} & 0.50 \\
\midrule

\multirow{2}{*}{\makecell[l]{\textbf{Reward Diversity}\\ 
\textbf{(Self-BLEU) \(\downarrow\)}}}
& RoboMoRe (w/o Diversity Reflection) & 0.70 & 0.67 & 0.70 & 0.57 & 0.70 & 0.75 & 0.74 & 0.72 \\
& \textbf{RoboMoRe (w/ Diversity Reflection)} & \textbf{0.50} & \textbf{0.46} & \textbf{0.46} & \textbf{0.48} & \textbf{0.47} & \textbf{0.45} & \textbf{0.43} & \textbf{0.42} \\
\bottomrule
\end{tabular}
}
\vspace{-.3cm}
\end{table*}



\paragraph{Diversity Reflection enables high quality and diverse morphology parameters and reward functions.}

Table \ref{tab:diversity_reflection_quantitative} illustrates that the Diversity Reflection mechanism substantially improves both the performance of generated samples and the diversity of rewards and morphologies. We compare our method against two baselines: (1) Random Sample (RS), a commonly used strategy in Evolutionary Algorithms (EA) and Bayesian optimization (BO), where initial designs are sampled from a Gaussian distribution \cite{frazier2018tutorial, back1993overview}, and (2) a setting without Diversity Reflection (w/o DR). Please refer to Appendix \ref{supp:Results for Diversity Reflection} for more details the efficiency comparison. Morphology diversity is quantified using the \textbf{coefficient of variation} \cite{bedeian2000use}, while reward diversity is assessed via the \textbf{Self-BLEU} metric \cite{zhu2018texygen}. All diversity metrics are averaged over fifty generated samples.

\begin{figure*}[ht]
  \centering
  \includegraphics[width=1.0\textwidth]{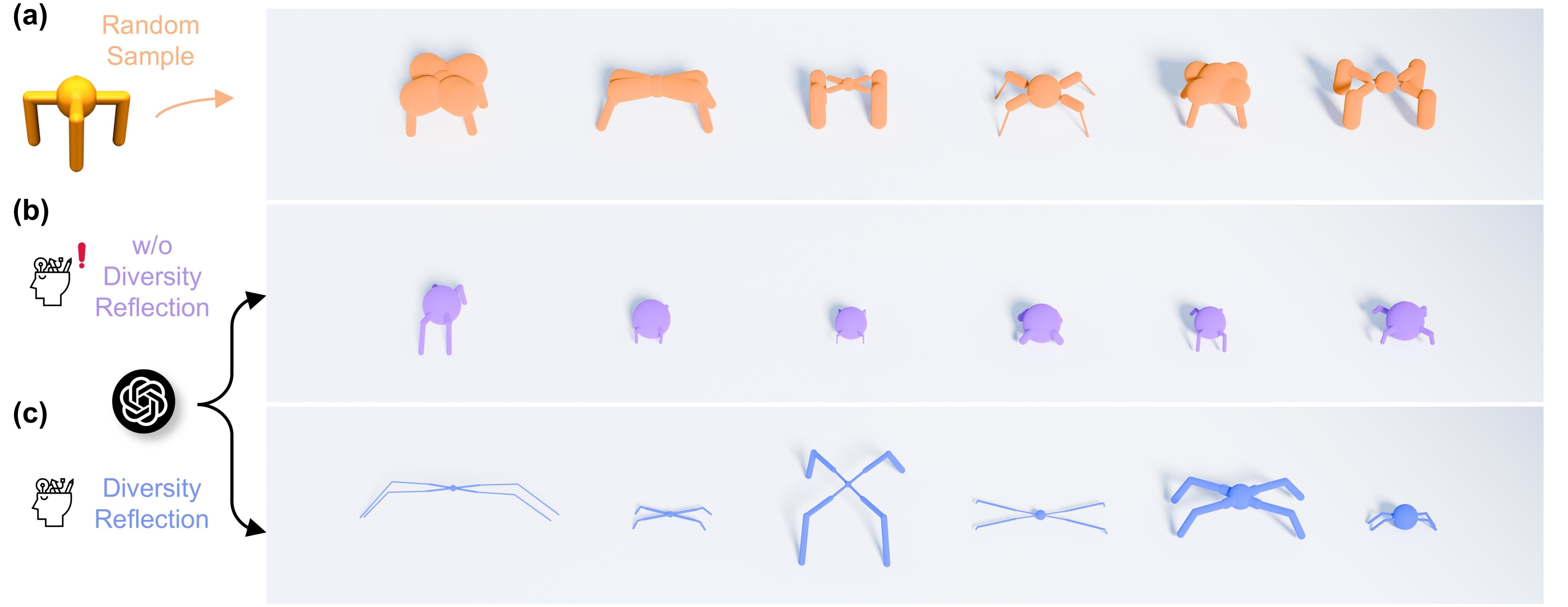}
  \caption{\textbf{Qualitative results using three different methods}: (a) Random Sample, (b) RoboMoRe (w/o Diversity Reflection), (c) RoboMoRe. It can be observed that RoboMoRe with Diversity Reflection could generate more high-quality and diverse robot morphology examples compared to others (see Table \ref{tab:diversity_reflection_quantitative}).}
  \label{fig:Fig_dr_comparison} 
  \vspace{-.3cm}
\end{figure*}
\begin{figure*}[t]
  \centering
  \includegraphics[width=1.0\textwidth]{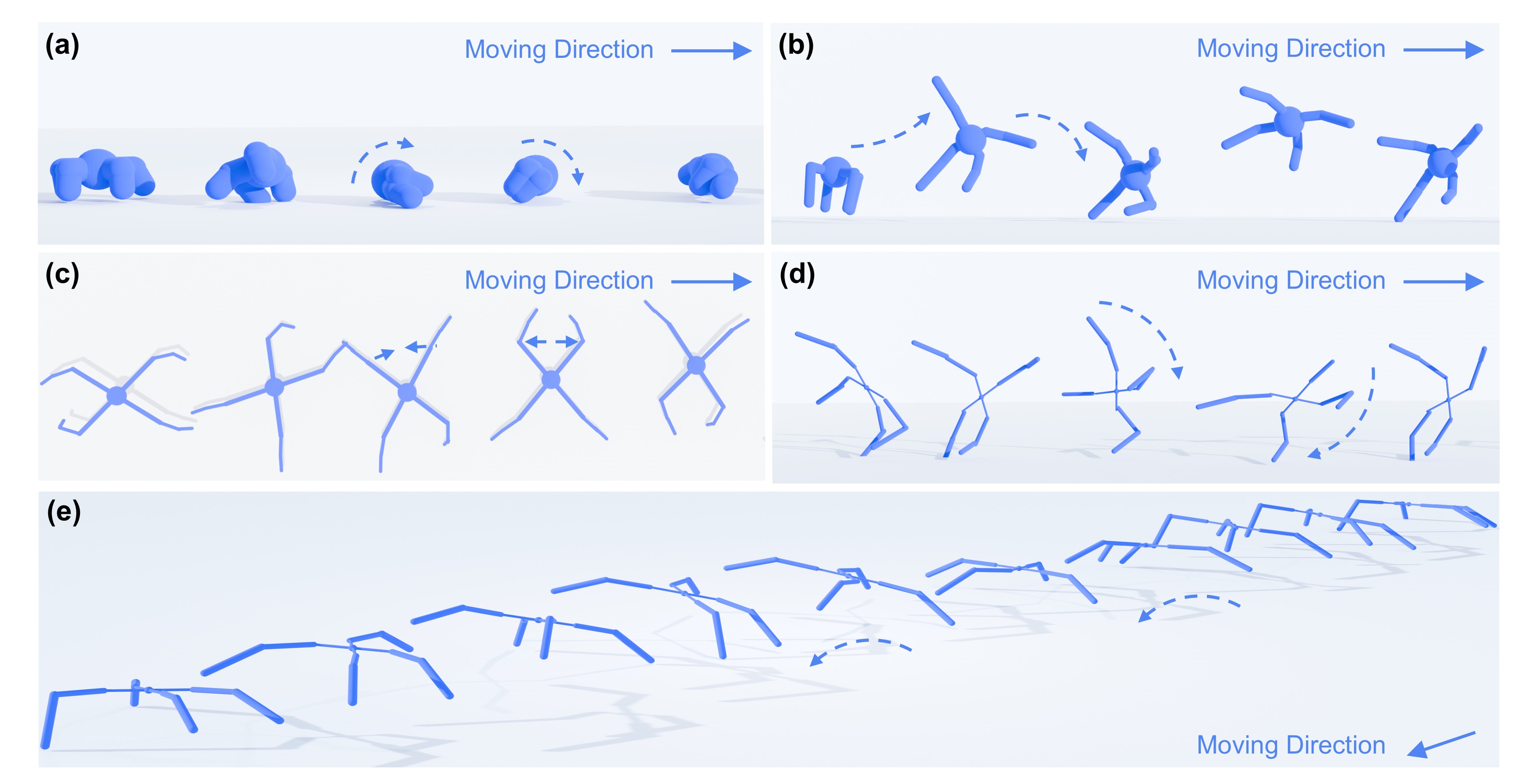} 
  \caption{\textbf{Qualitative results of diverse motion behaviors using Diversity Reflection.} (a) The short-legged, low-center-of-gravity robot rolls forward along the ground. (b) The uniform robot propels itself forward with sideways hops. (c) The crab-inspired robot repositions itself and scuttles forward in a crab-walking manner. (d) The agile, lightweight robot advances by executing a side flip. (e) The agile, long-legged robot moves forward with a light hopping gait.}
  \label{fig:Fig_diversity_morrew} 
\vspace{-.3cm}
\end{figure*}

While Random Sample introduces diversity, it often generates low-quality designs—such as bulky and unbalanced legs for the Ant robot, see Fig. \ref{fig:Fig_dr_comparison}. This low initial quality explains why traditional methods such as EA and BO typically require thousands of iterations to converge. In contrast, LLMs, even without diversity reflection, are capable of generating higher-quality samples thanks to their inherent reasoning and design capabilities. However, without explicitly promoting diversity, such methods fail to explore more performant regions of the design space, resulting in suboptimal performance. RoboMoRe achieves the best overall performance because it effectively leverages Diversity Reflection to generate samples that are both high-quality and diverse, enabling a more efficient and robust design optimization process. Fig. \ref{fig:Fig_diversity_morrew} provides more qualitative results for morphologies and rewards. 

\subsection{Effectiveness of Reward Shaping and Morphology Design}
\label{sec:reward_shaping}

\paragraph{Reward Shaping can improve performance by inspiring learning morphology-suited motion behaviors.}
Table \ref{tab:robodesign_ablation} demonstrates that reward shaping improves performance across most tasks. This is primarily because different robot designs tend to favor different optimal locomotion strategies, and reward shaping enables the discovery of such task-specific optimal behaviors. Notably, our findings indicate that reward shaping yields greater performance gains in tasks characterized by richer morphological parameterization and more complex state spaces. For instance, Half-Cheetah, which has more than 20 morphological parameters and is the most complex design among the evaluated robots benefits the most. In contrast, the simplest robot, Swimmer—characterized by a minimal set of 6 design parameters and a constrained, two-dimensional action space—demonstrates only marginal benefits from reward shaping. 

\begin{table*}[ht]
\centering
\caption{\textbf{Ablation study of RoboDesign across efficiency metrics under different Reward (Rew.) and Morphology (Mor.) configurations.}}
\label{tab:robodesign_ablation}
\resizebox{\textwidth}{!}{
\begin{tabular}{llrrrrrrrr}
\toprule
\textbf{Method} & \textbf{Metric} & \textbf{Ant} & \textbf{Ant-Powered}& \textbf{Ant-Desert} & \textbf{Ant-Jump} & \textbf{Hopper} & \textbf{Half-Cheetah} & \textbf{Swimmer} & \textbf{Walker} \\
\midrule

\multirow{4}{*}{Efficiency \(\uparrow\)} 
& RoboDesign (w/o Mor. \& Rew.) & 21.64 & 8.68 & 20.69 & 6.08 & 94.19 & 4,984.63 & 6.82 & 15.36 \\
& RoboDesign (w/o Mor.)        & 97.74 & 18.04 & 15.79 & 279.42 & 561.61 & 7,945.81 & 15.63 & 22.37 \\
& RoboDesign (w/o Rew.)        & 6,190.39 & 3,026.53 & 3,742.68 & 6.34 & 1,653.59 & 67,992.31 & 22,009.94 & 1401.21 \\
& \textbf{RoboDesign (Full)}   & \textbf{10,464.17} & \textbf{6,537.45} & \textbf{8,001.67} & \textbf{396.45} & \textbf{1,951.44} & \textbf{129,440.98} & \textbf{21,931.15} & \textbf{1,482.60} \\
\bottomrule
\end{tabular}
}
\vspace{-0.5cm}
\end{table*}

\paragraph{Morphology Design still plays a fundamental role.}
Although reward shaping can significantly enhance performance by promoting the emergence of novel motion patterns, morphology design remains crucial. When a fixed, template-based morphology is used, performance degrades markedly across all tasks, suggesting that human-designed morphologies are still far from optimal—consistent. Moreover, we observe substantial variation in material usage across different morphology designs, which accounts for the pronounced disparities in efficiency observed in our experiments.

\subsection{Effectiveness of Coarse-to-Fine Optimization.}
\begin{wrapfigure}{r}{0.45\textwidth}
  \centering
  \includegraphics[width=0.5\textwidth]{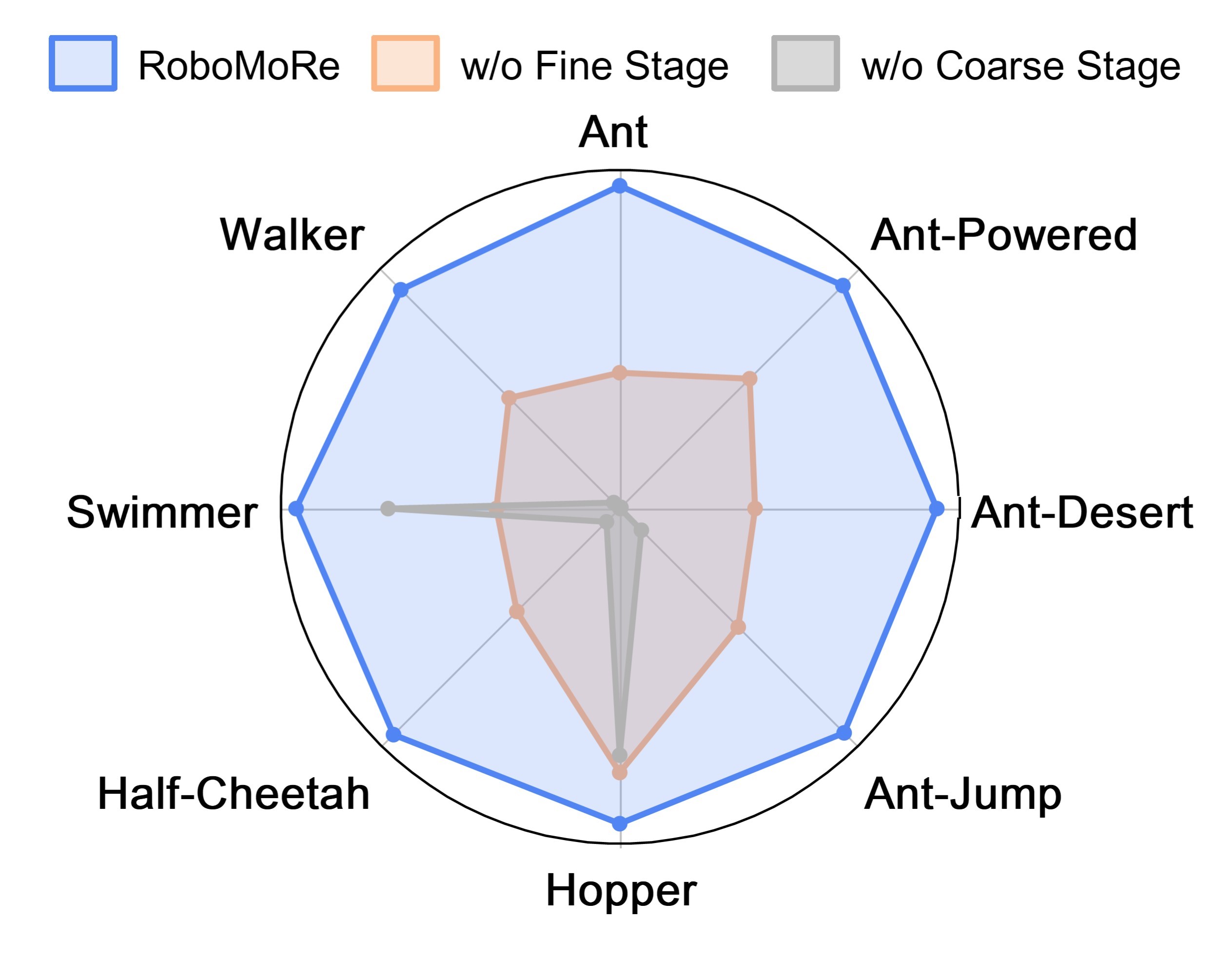} 
  \caption{\textbf{Comparison of the highest efficiency achieved in different cases among RoboMoRe, RoboMoRe (w/o fine stage), and RoboMoRe (w/o coarse stage).}}
    \label{fig:coarsetofine}

\end{wrapfigure}
    \textbf{Both coarse and fine stages contribute significantly to co-design performance.} Fig.~\ref{fig:coarsetofine} highlights the critical role of jointly leveraging both coarse and fine stages in achieving optimal co-design performance. Across all tasks, the fine stage consistently enhances the results of the coarse stage. We attribute this to the greater design space inherent in these more challenging tasks, which, while offering more room for optimization, also make it significantly harder for LLMs to directly generate near-optimal designs. The coarse stage enables broad exploration of the joint morphology–reward space, generating diverse, high-quality candidates. However, in complex or high-dimensional tasks, these alone are insufficient. The fine stage complements by alternately refining morphology and reward using gradient-based feedback. Its effectiveness crucially depends on the search space shaped by the coarse stage. The fine stage alone lags behind the full approach in most cases, but matches or outperforms the coarse stage in simpler environments like Swimmer and Hopper.

\section{Discussion \& Conclusion}

RoboMoRe is designed to be task-agnostic and broadly generalizable, but future work should test its robustness on more challenging benchmarks, such as soft robotics platforms \cite{graule2022somogym, gazzola2018forward, zhang2019modeling}, where training a single robot can take over a day, making large-scale optimization significantly harder. We also observe that well-performing morphologies under the template reward often remain efficient, while poor ones rarely improve—suggesting early-stage pruning as a viable strategy (Appendix \ref{supp:Results for Diversity Reflection}). Incorporating vision-language models (VLMs) may further enhance adaptability by mimicking human design intuition. Finally, we aim to extend RoboMoRe to real-world tasks, enabling bio-inspired robots to co-optimize morphology and reward for more efficient motion.

In conclusion, we presented RoboMoRe, an LLM-driven framework that integrates reward shaping into the robot co-design process. By adopting a coarse-to-fine optimization strategy—combining Diversity Reflection with alternating LLM-guided refinement—RoboMoRe successfully discovers efficient and high-performing morphology-reward pairs. Across eight robot design tasks, our method consistently outperforms human baselines and prior co-design approaches. We hope RoboMoRe contributes to a deeper understanding of joint optimization in robot design and serves as a springboard for future innovations at the intersection of generative AI and embodied intelligence.


\clearpage
\newpage
\bibliography{main}

\newpage
\appendix

\section*{Appendix}

\section{Full Prompts}
\label{supp:prompts}
\subsection{Coarse Optimization Prompts}

We take the Ant as an example to illustrate the prompt used in coarse optimization. The prompt for morphology design mainly consists of three parts: the task description, a masked MuJoCo XML file, and the required output format. The MuJoCo XML is masked to prevent LLMs from seeing the human-designed structure in advance. For brevity, we show only part of the MuJoCo XML file and environment code. Please refer to MuJoCo Gym for more details \cite{towers2024gymnasium}.
\subsubsection{Prompts for Reward Shaping}
\begin{lstlisting}[style=pythonstyle,label={task_description}]
You are a reward engineer trying to write reward functions to solve reinforcement learning tasks as effectively as possible.
Your goal is to write a reward function for the enviroment that will help the agent learn the task described in text.

Task Description: The ant is a 3D quadruped robot consisting of a torso (free rotational body) with four legs attached to it, where each leg has two body parts. 
The goal is to coordinate the four legs to move in the forward (right) direction by applying torque to the eight hinges connecting the two body parts of each leg and the torso (nine body parts and eight hinges), You should write a reward function to make the robot move as faster as possible.

Here is the environment codes:
class AntEnv(MujocoEnv, utils.EzPickle):
<Enviroment Code>

A template reward can be:
    def _get_rew(self, x_velocity: float, action):
        <reward function code you should write>
        return reward, reward_info    

The output of the reward function should consist of two items:  
(1) `reward`, which is the total reward.  
(2) `reward_info`, a dictionary of each individual reward component.  

The code output should be formatted as a Python code string: `'```python ...```'`.  

Some helpful tips for writing the reward function code:  
(1) You may find it helpful to normalize the reward to a fixed range by applying transformations like `numpy.exp` to the overall reward or its components.    
(2) Make sure the type of each input variable is correctly specified and the function name is "def _get_rew():"
(3) Most importantly, the reward code's input variables must contain only attributes of the provided environment class definition (namely, variables that have the prefix `self.`). Under no circumstances can you introduce new input variables.


\end{lstlisting}



\subsubsection{Prompts for Morphology Design}
\begin{lstlisting}[style=pythonstyle,label={1st:task_description}]
Role: You are a robot designer trying to design robot parameters to increase the fitness function as effective as possible.
Your goal is to design parameters of robot that will help agent achieve the fitness function as high as possible.
Fintess function: walk distance/material cost.
Task Description: The ant is a 3D quadruped robot consisting of a torso (free rotational body) with four legs attached to it, where each leg has two body parts. The goal is to coordinate the four legs to move in the forward (right) direction by applying torque to the eight hinges connecting the two body parts of each leg and the torso (nine body parts and eight hinges).
Here is the xml file:
"""
<mujoco model="ant">
...
  <worldbody>
    <light cutoff="100" diffuse="1 1 1" dir="-0 0 -1.3" directional="true" exponent="1" pos="0 0 1.3" specular=".1 .1 .1"/>
    <geom conaffinity="1" condim="3" material="MatPlane" name="floor" pos="0 0 0" rgba="0.8 0.9 0.8 1" size="40 40 40" type="plane"/>
    <body name="torso" pos="0 0 {height}">
      <camera name="track" mode="trackcom" pos="0 -3 0.3" xyaxes="1 0 0 0 0 1"/>
      <geom name="torso_geom" pos="0 0 0" size="{param1}" type="sphere"/>
      <joint armature="0" damping="0" limited="false" margin="0.01" name="root" pos="0 0 0" type="free"/>
      <body name="front_left_leg" pos="0 0 0">
        <geom fromto="0.0 0.0 0.0 {param2} {param3} 0.0" name="aux_1_geom" size="{param8}" type="capsule"/>
        <body name="aux_1" pos="{param2} {param3} 0">
          <joint axis="0 0 1" name="hip_1" pos="0.0 0.0 0.0" range="-40 40" type="hinge"/>
          <geom fromto="0.0 0.0 0.0 {param4} {param5} 0.0" name="left_leg_geom" size="{param9}" type="capsule" />
          <body pos="{param4} {param5} 0" >
            <joint axis="-1 1 0" name="ankle_1" pos="0.0 0.0 0.0" range="30 100" type="hinge"/>
            <geom fromto="0.0 0.0 0.0 {param6} {param7} 0.0" name="left_ankle_geom" size="{param10}" type="capsule"/>
          </body>
        </body>
      </body>
...
"""
</mujoco>
Attention:
1. For reducing material cost to ensure the effieciency of robot design, you should reduce redundant paramters and increase paramters who control the robot.
2. Your design should fit the control gear and others parts of robots well.
Output Format:
Please output in json format without any notes:
{ "parameters": [<param1>, <param2>, ..., <param10>], 
"description": "<your simple design style description>", } 
#   param1 is the size of the torso, positive.
#   param2 is is the end attachment x point of leg, positive. 
#   param3 is is the end attachment y point of leg, positive. 
#   param4 is is the end attachment x point of hip, positive. 
#   param5 is the end attachment y point of hip, positive. 
#   param6 is the end attachment x point of ankle, positive.
#   param7 is the end attachment y point of ankle, positive.
#   param8 is the size of the leg, positive.
#   param9 is the size of the hip, positive.
#   param10 is the size of the ankle, positive.
\end{lstlisting}



\subsubsection{Prompts for Diversity Reflection}
\begin{lstlisting}[style=pythonstyle,label={1st:task_description}]
Please propose a new morphology design, which can promote high fitness function and is quite different from all previous morphology designs in the design style. (Morphology)
\end{lstlisting}

\begin{lstlisting}[style=pythonstyle,label={1st:task_description}]
Please write a new reward function to encourage more robot motion behaviours, which can promote high fitness function and is quite different from all previous reward functions in the design style. (Reward)
\end{lstlisting}


\subsection{Fine Optimization}

\subsubsection{Prompts for Alternating Refinement}
\begin{lstlisting}[style=pythonstyle,label={1st:task_description}]
There are also some design parameters and their fitness. Please carefully observe these parameters and their fitness, try to design a new parameter to further improve the fitness. (Morphology)
\end{lstlisting}

\begin{lstlisting}[style=pythonstyle,label={1st:task_description}]
There are also some reward functions and their fitness. Please carefully observe these reward funcions and their fitness, try to write a reward function to further improve the fitness. (Reward)
\end{lstlisting}


\clearpage
\newpage

\section{Motivation}
\label{supp:motivation}
To verify our hypothesis, we conducted experiments to quantitatively evaluate the performance of two representative robot morphologies, short leg ant and long leg ant. By designing different reward functions, we encouraged each robot to adopt distinct motion modalities. Specifically, we encouraged the robot to adopt a rolling motion along the ground by augmenting the template reward function with an angular velocity term and a jumping motion on the air by augmenting the template reward function with a bouncing reward term.

\begin{figure*}[htbp]
  \centering
  \includegraphics[width=1.0\textwidth]{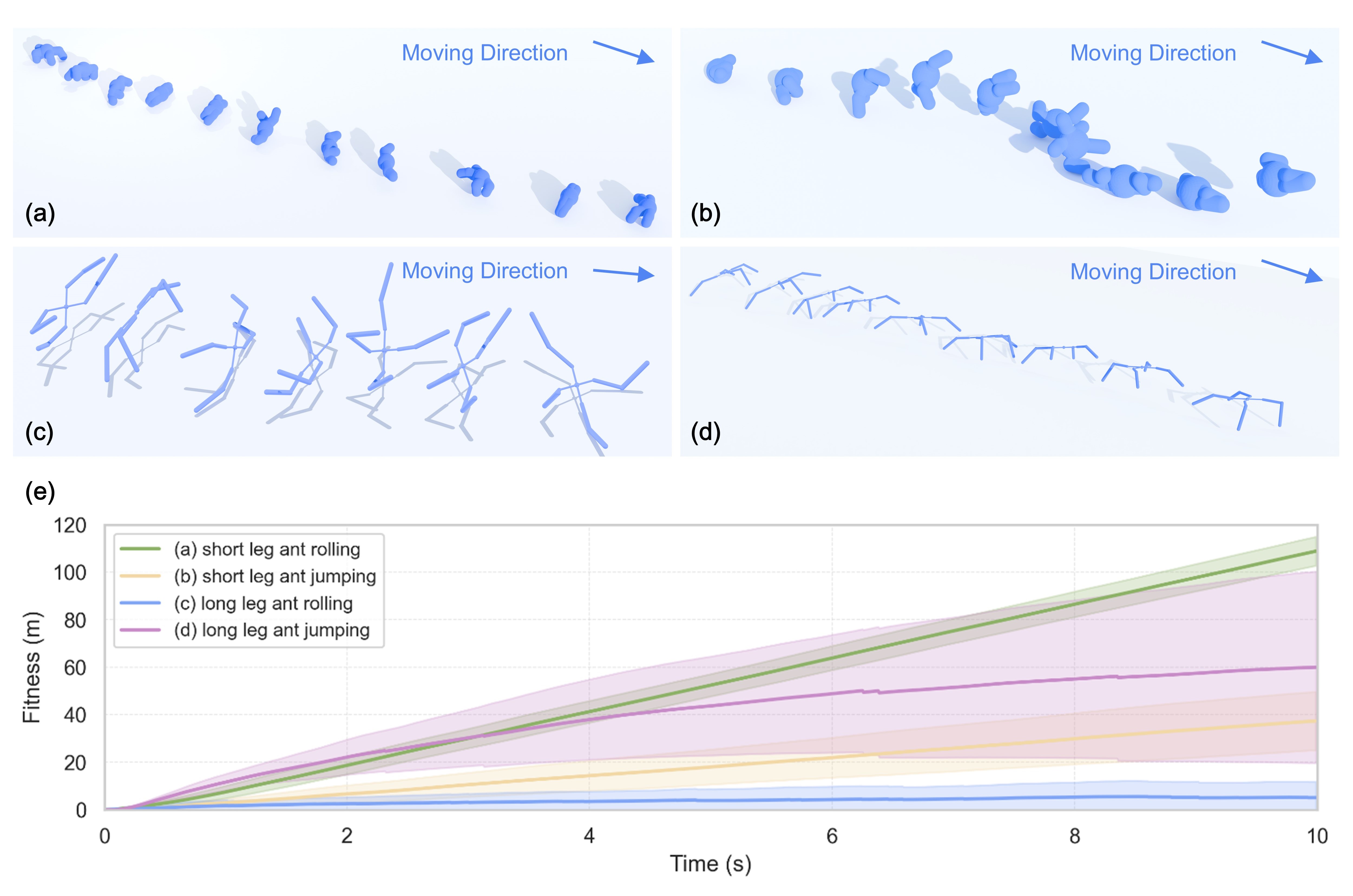} 
  \caption{\textbf{(a-d) Motion behaviors of our four cases. (e) Time-series fitness (walk distance) across 100 independent evaluations per design case.}}
  \label{fig:motivation} 
\end{figure*}

As shown in Fig. \ref{fig:motivation}, the experimental results indicate that the short leg ant, characterized by a bulky body and low center of gravity, exhibits low fitness in the jumping modality but achieves significantly higher fitness when utilizing the rolling modality. Conversely, the long leg ant robot, benefiting from its slender limbs and elevated center of gravity, performs exceptionally well in the jumping modality. However, its motion efficiency drastically decreases under the rolling modality due to frequent overturning and difficulty recovering balance. These results clearly illustrate the strong dependence of robot performance
on appropriate motion modalities, highlighting that optimal efficiency can only be achieved when
morphology and modality are suitably matched.

\begin{wrapfigure}{r}{0.5\linewidth}
\vspace{-3mm}  
\centering
\captionof{table}{\textbf{Fitness and Efficiency on four cases.}}
\label{tab:half_right}
\begin{tabular}{lcccc}
\toprule
\textbf{Case} & (a) & (b) & (c) & (d) \\
\midrule
Fitness & 108.83 & 37.39 & 59.94 & 5.13\\
Efficiency & 288.76  & 109.50  & 1457.65 & 124.75 \\
\bottomrule
\end{tabular}
\vspace{-2mm}  
\end{wrapfigure}

In addition, Table \ref{tab:half_right} presents the fitness and efficiency of the ant robot under different conditions. We observe that the long-leg ant exhibits a clear advantage in terms of efficiency, but only when paired with a well-suited reward function. This finding underscores the importance of co-optimizing both morphology and control strategy in robot design to achieve optimal efficiency.

\clearpage
\newpage

\section{Environment and Task Details}
This section provides additional details on the environments and tasks used in our experiments. We build upon standard environments from MuJoCo Gym, including Ant, Walker, Hopper, Half-Cheetah, and Swimmer, and use their built-in reward functions and morphologies as human-designed baselines for comparison. To further evaluate the robustness of RoboMoRe across diverse settings, we introduce three customized environments: Ant-Desert, Ant-Jump, and Ant-Powered. Task descriptions are used as input to both the Morphology Design Prompt and the Reward Shaping Prompt. The output format constrains the LLM's response to a set of robot morphology parameters, which are then fed into our custom design scripts to generate a complete MuJoCo XML file, following \cite{ha2019reinforcement}. We will open-source these design scripts alongside the code.

\label{supp:env}

\begin{figure*}[htbp]
  \centering
  \includegraphics[width=1.0\textwidth]{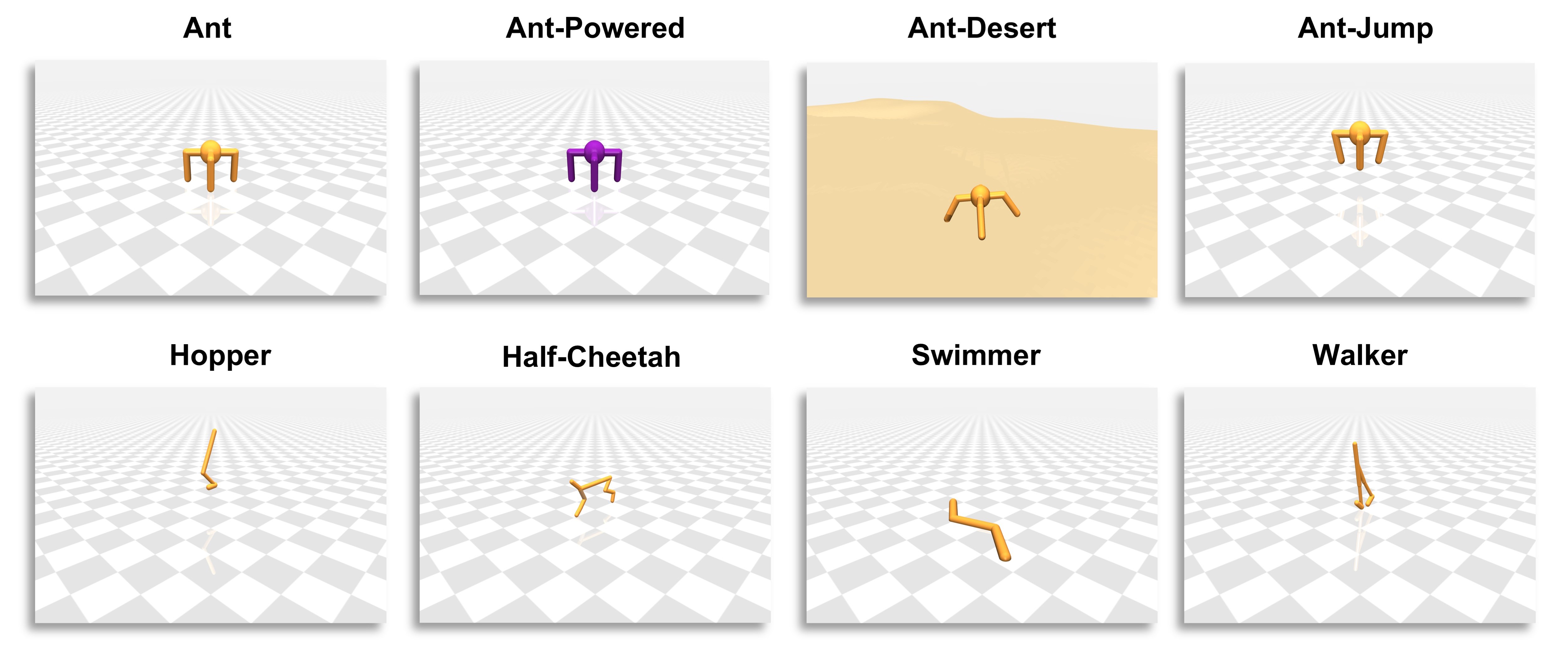} 
  \caption{\textbf{Human designed agents in eight different environments for visualization.} Purple
indicates agents in an augmented gear power.}
  \label{fig:environment} 
\end{figure*}

\subsection{Ant}
\subsubsection{Task Description}
\begin{lstlisting}[style=pythonstyle,label={1st:task_description}]
The ant is a 3D quadruped robot consisting of a torso (free rotational body) with four legs attached to
it, where each leg has two body parts. The goal is to coordinate the four legs to move in the forward
(right) direction by applying torque to the eight hinges connecting the two body parts of each leg and
the torso (nine body parts and eight hinges).
\end{lstlisting}
\subsubsection{Output Format}
\begin{lstlisting}[style=pythonstyle,label={1st:task_description}]
Please output in json format without any notes:
{ "parameters": [<param1>, <param2>, ..., <param10>], 
"description": "<your simple design style description>", } 
#   param1 is the size of the torso, positive.
#   param2 is is the end attachment x point of leg, positive. 
#   param3 is is the eend attachment y point of leg, positive. 
#   param4 is is the end attachment x point of hip, positive. 
#   param5 is the end attachment y point of hip, positive. 
#   param6 is the end attachment x point of ankle, positive.
#   param7 is the end attachment y point of ankle, positive.
#   param8 is the size of the leg, positive.
#   param9 is the size of the hip, positive.
#   param10 is the size of the ankle, positive.
\end{lstlisting}

\subsection{Ant-Powered}
\subsubsection{Task Description}
\begin{lstlisting}[style=pythonstyle,label={1st:task_description}]
The ant is a 3D quadruped robot consisting of a torso (free rotational body) with four legs attached to it, where each leg has two body parts. The goal is to coordinate the four legs to move in the forward (right) direction by applying torque to the eight hinges connecting the two body parts of each leg and the torso (nine body parts and eight hinges)
\end{lstlisting}
\subsubsection{Output Format}
\begin{lstlisting}[style=pythonstyle,label={1st:task_description}]
Please output in json format without any notes:
{ "parameters": [<param1>, <param2>, ..., <param10>], 
"description": "<your simple design style description>", } 
#   param1 is the size of the torso, positive.
#   param2 is is the end attachment x point of leg, positive. 
#   param3 is is the end attachment y point of leg, positive. 
#   param4 is is the end attachment x point of hip, positive. 
#   param5 is the end attachment y point of hip, positive. 
#   param6 is the end attachment x point of ankle, positive.
#   param7 is the end attachment y point of ankle, positive.
#   param8 is the size of the leg, positive.
#   param9 is the size of the hip, positive.
#   param10 is the size of the ankle, positive.
\end{lstlisting}

\subsection{Ant-Desert}
\subsubsection{Task Description}
\begin{lstlisting}[style=pythonstyle,label={1st:task_description}]
The ant is a 3D quadruped robot consisting of a torso (free rotational body) with four legs attached to it, where each leg has two body parts. The goal is to coordinate the four legs to move in the forward (right) direction by applying torque to the eight hinges connecting the two body parts of each leg and the torso (nine body parts and eight hinges).
\end{lstlisting}
\subsubsection{Output Format}
\begin{lstlisting}[style=pythonstyle,label={1st:task_description}]
Please output in json format without any notes:
{ "parameters": [<param1>, <param2>, ..., <param10>], 
"description": "<your simple design style description>", } 
#   param1 is the size of the torso, positive.
#   param2 is is the end attachment x point of leg, positive. 
#   param3 is is the end attachment y point of leg, positive. 
#   param4 is is the end attachment x point of hip, positive. 
#   param5 is the end attachment y point of hip, positive. 
#   param6 is the end attachment x point of ankle, positive.
#   param7 is the end attachment y point of ankle, positive.
#   param8 is the size of the leg, positive.
#   param9 is the size of the hip, positive.
#   param10 is the size of the ankle, positive.
\end{lstlisting}

\subsection{Ant-Jump} 
\subsubsection{Task Description}
\begin{lstlisting}[style=pythonstyle,label={1st:task_description}]
The ant is a 3D quadruped robot consisting of a torso (free rotational body) with four legs attached to it, where each leg has two body parts. The goal is to coordinate the four legs to jump in the upward direction by applying torque to the eight hinges connecting the two body parts of each leg and the torso (nine body parts and eight hinges)
\end{lstlisting}
\subsubsection{Output Format}
\begin{lstlisting}[style=pythonstyle,label={1st:task_description}]
Please output in json format without any notes:
{ "parameters": [<param1>, <param2>, ..., <param10>], 
"description": "<your simple design style description>", } 
#   param1 is the size of the torso, positive.
#   param2 is is the end attachment x point of leg, positive. 
#   param3 is is the end attachment y point of leg, positive. 
#   param4 is is the end attachment x point of hip, positive. 
#   param5 is the end attachment y point of hip, positive. 
#   param6 is the end attachment x point of ankle, positive.
#   param7 is the end attachment y point of ankle, positive.
#   param8 is the size of the leg, positive.
#   param9 is the size of the hip, positive.
#   param10 is the size of the ankle, positive.
\end{lstlisting}

\subsection{Hopper}
\subsubsection{Task Description}
\begin{lstlisting}[style=pythonstyle,label={1st:task_description}]
The hopper is a two-dimensional one-legged figure consisting of four main body parts - the torso at the top, the thigh in the middle, the leg at the bottom, and a single foot on which the entire body rests. The goal is to make hops that move in the forward (right) direction by applying torque to the three hinges that connect the four body parts.
\end{lstlisting}
\subsubsection{Output Format}
\begin{lstlisting}[style=pythonstyle,label={1st:task_description}]
Please output in json format without any notes:
{ "parameters": [<param1>, <param2>, ..., <param10>], 
"description": "<your simple design style description>", } 
#   param1 is the size of the torso, positive.
#   param2 is is the end attachment x point of leg, positive. 
#   param3 is is the end attachment y point of leg, positive. 
#   param4 is is the end attachment x point of hip, positive. 
#   param5 is the end attachment y point of hip, positive. 
#   param6 is the end attachment x point of ankle, positive.
#   param7 is the end attachment y point of ankle, positive.
#   param8 is the size of the leg, positive.
#   param9 is the size of the hip, positive.
#   param10 is the size of the ankle, positive.
\end{lstlisting}

\subsection{Half-Cheetah}
\begin{lstlisting}[style=pythonstyle,label={1st:task_description}]
The HalfCheetah is a 2-dimensional robot consisting of 9 body parts and 8 joints connecting them (including two paws). The goal is to apply torque to the joints to make the cheetah run forward (right) as fast as possible, with a positive reward based on the distance moved forward and a negative reward for moving backward.
The cheetah's torso and head are fixed, and torque can only be applied to the other 6 joints over the front and back thighs (which connect to the torso), the shins (which connect to the thighs), and the feet (which connect to the shins), and the goal is to move as fast as possible towards the right by applying torque to the rotors and using fluid friction.
\end{lstlisting}
\subsubsection{Output Format}
\begin{lstlisting}[style=pythonstyle,label={1st:task_description}]
Please output in json format without any notes:
{"parameters": [<param1>, <param2>, ..., <param24>],
"description": "<your simple design style decription>",}
Parameters Description:
# param1: start point of torso (x), negtive
# param2: end point of torso (x) and start point of head (x), positive
# param3: end point of head (x).
# param4: end point of head (z).
# back leg parameters
# param5: end point of bthigh (x) and start point of bshin (x), sometimes positive.
# param6: end point of bthigh (z) and start point of bshin (z), negative.
# param7: end point of bshin (x) and start point of bfoot (x), sometimes negtive.
# param8: end point of bshin (z) and start point of bfoot (z), negative.
# param9: end point of bfoot (x), sometimes positive.
# param10: end point of bfoot (z), negative.
# forward leg parameters
# param11: end point of fthigh (x) and start point of fshin (x), sometimes negative.
# param12: end point of fthigh (z) and start point of fshin (z), negative.
# param13: end point of fshin (x) and start point of ffoot (x), sometimes positive.
# param14: end point of fshin (z) and start point of ffoot (z), negative.
# param15: end point of ffoot (x), sometimes positive.
# param16: end point of ffoot (z), negative.
# size information
# param17: torso capsule size.
# param18: head capsule size.
# param19: bthigh capsule size.
# param20: bshin capsule size.
# param21: bfoot capsule size.
# param22: fthigh capsule size.
# param23: fshin capsule size.
# param24: ffoot capsule size.
\end{lstlisting}

\subsection{Swimmer}
\begin{lstlisting}[style=pythonstyle,label={1st:task_description}]
The swimmers consist of three or more segments ('links') and one less articulation joints ('rotors') - one rotor joint connects exactly two links to form a linear chain. The swimmer is suspended in a two-dimensional pool and always starts in the same position (subject to some deviation drawn from a uniform distribution), and the goal is to move as fast as possible towards the right by applying torque to the rotors and using fluid friction.
\end{lstlisting}
\subsubsection{Output Format}
\begin{lstlisting}[style=pythonstyle,label={1st:task_description}]
Please output in json format without any notes:
{"parameters": [<param1>, <param2>, ..., <param6>],
"description": "<your simple design style decription>",}
Parameters Description:
#   param1 is the length of first segment, positive.
#   param2 is the length of second segment, positive.
#   param3 is the length of third segment, positive.
#   param4 is the size of first segment, positive.
#   param5 is the size of second segment, positive.
#   param6 is the size of third segment, positive.
\end{lstlisting}

\subsection{Walker}
\subsubsection{Task Description}
\begin{lstlisting}[style=pythonstyle,label={1st:task_description}]
The walker is a two-dimensional bipedal robot consisting of seven main body parts - a single torso at the top (with the two legs splitting after the torso), two thighs in the middle below the torso, two legs below the thighs, and two feet attached to the legs on which the entire body rests. The goal is to walk in the forward (right) direction by applying torque to the six hinges connecting the seven body parts.
\end{lstlisting}
\subsubsection{Output Format}
\begin{lstlisting}[style=pythonstyle,label={1st:task_description}]
Please output in json format without any notes:
{ "parameters": [<param1>, <param2>, ..., <param10>], 
"description": "<your simple design style description>", } 
#   param1 is the size of the torso, positive.
#   param2 is is the end attachment x point of leg, positive. 
#   param3 is is the end attachment y point of leg, positive. 
#   param4 is is the end attachment x point of hip, positive. 
#   param5 is the end attachment y point of hip, positive. 
#   param6 is the end attachment x point of ankle, positive.
#   param7 is the end attachment y point of ankle, positive.
#   param8 is the size of the leg, positive.
#   param9 is the size of the hip, positive.
#   param10 is the size of the ankle, positive.
\end{lstlisting}

\newpage

\section{Implementation Details}
\label{supp:Implementation Details}
\begin{wraptable}{r}{0.5\textwidth}
\centering
\caption{\textbf{Coarse-to-Fine Optimization Hyperparameters}}
\label{tab:coarse_fine_hyperparams}
\begin{tabular}{lc}
\hline
\textbf{Hyperparameter} & \textbf{Value} \\
\hline
$\mathcal{N_M}$ & 25 \\
$\mathcal{N_R}$ & 5 \\
$k$ & 5\% \\
\hline
\end{tabular}
\end{wraptable}
In line with standard reinforcement learning practices, we employ distributed trajectory sampling across multiple CPU threads to accelerate training. Each model is trained using four random seeds on a system equipped with one AMD EPYC 7T83 processor and a single NVIDIA RTX 4080 Super GPU. Our framework is built on Python 3.8.20, MuJoCo 2.3.1, Stable-Baselines3 2.4.1, and CUDA 12.4. For all environments considered, training a single policy for 5e5 steps takes approximately 15 minutes using 24 CPU cores and one RTX 4080 Super GPU. GPT-4-turbo serves as the foundation model in most of our experiments. As shown in Table \ref{tab:coarse_fine_hyperparams}, for coarse optimization, we empirically set the number of morphologies $\mathcal{N_M}$ to 25 and the number of reward functions $\mathcal{N_R}$ to 5, for trade-off between speed and design quality. For fine optimization, we select $k$ as 5\% best candidates from coarse stage.  Although this process involves 120 iterations, it remains computationally efficient and practical—especially when compared to conventional methods like Bayesian Optimization or Evolutionary Algorithms, which often require several thousand iterations to converge.

\paragraph{Policy Training} 

We use the Soft Actor-Critic (SAC) algorithm \cite{haarnoja2018soft} from Stable-Baselines3 as the reinforcement learning backbone for all experiments, ensuring consistency across tasks. Hyperparameters are configured based on the recommendations from Gym SpinningUp \cite{SpinningUp2018}, which are well-suited for training both human-designed robot morphologies and their corresponding reward functions. To accelerate training, we leverage parallel environments via SubprocVecEnv with 16 instances. Experimental results are averaged over 100 independent runs to mitigate randomness. Each candidate is initially trained for \(5 \times 10^5\) steps for efficiency, and the top-performing candidate is retrained for \(10^6\) steps to ensure a fair comparison with alternative methods such as Bayesian Optimization (BO) and Eureka—both of which typically converge within this training budget across most tasks \cite{SpinningUp2018}. Table \ref{tab:sac_params} demonstrates detailed parameters for policy training.

\begin{table}[ht]
\centering
\caption{\textbf{SAC Training Hyperparameters}}
\label{tab:sac_params}
\begin{tabular}{ll}
\hline
\textbf{Parameter} & \textbf{Value} \\
\hline
Number of environments & 16 \\
Learning rate & 3e-4 \\
Buffer size & 2,000,000 \\
Learning starts & 10,000 \\
Batch size & 1024 \\
$\tau$ & 0.005 \\
$\gamma$ & 0.99 \\
Train frequency & 8 \\
Gradient steps & 4 \\
Policy kwargs & {[}512, 512{]} \\
\hline
\end{tabular}
\end{table}

\paragraph{Comparison Methods} We adopt Bayesian Optimization with a batch size of 1 and 100 iterations, using a Gaussian Process surrogate model with a Matérn 5/2 kernel, automatic relevance determination (ARD), and the Expected Improvement (EI) acquisition function. For Eureka, we follow the recommended configuration with 5 iterations and 16 populations per iteration.

\newpage

\section{Additional Results}
\subsection{Results for Diversity Reflection}
\label{supp:Results for Diversity Reflection}
\begin{figure*}[htbp]
  \centering
  \includegraphics[width=1.0\textwidth]{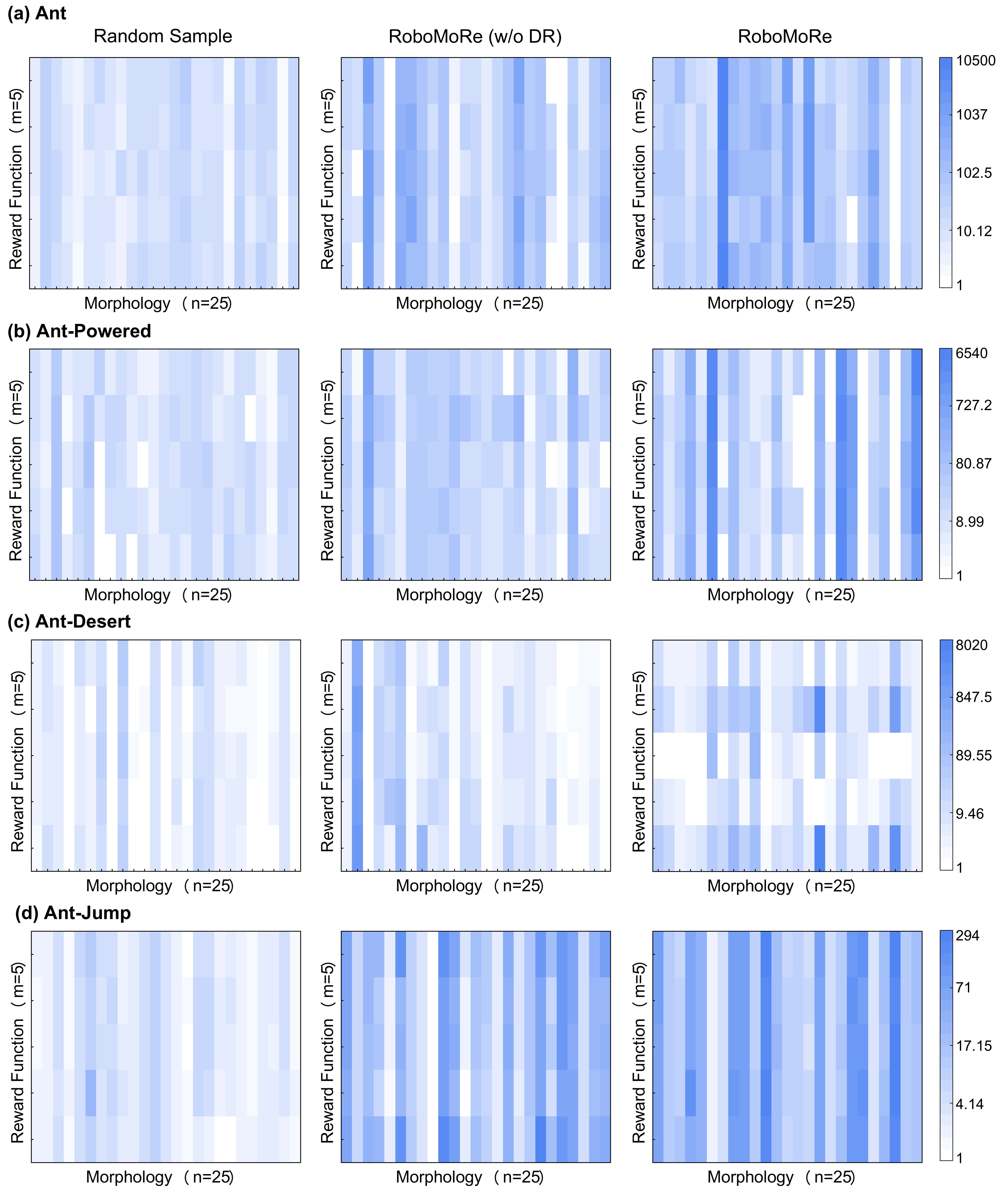} 
  \label{Fig:CoarseA_E} 
\end{figure*}

To more concretely demonstrate the effectiveness of the Diversity Reflection mechanism, we visualize the 5 × 25 efficiency matrices across multiple tasks for Random Sample, RoboMoRe w/o Diversity Reflection (DR), and RoboMoRe. Due to the large variation in data magnitude, we apply a logarithmic scale to the color map. As shown in Fig. \ref{Fig:CoarseA_E}, robots generated by Random Sample generally exhibit low efficiency, whereas those designed by RoboMoRe consistently achieve significantly higher performance across most tasks (see Table \ref{tab:diversity_reflection_quantitative}). Notably, while RoboMoRe w/o Dr still achieves reasonably good results, the lack of morphological diversity causes the selected samples to concentrate around local optima, thereby leading to a slightly lower average efficiency compared to the full RoboMoRe configuration.

\begin{figure*}[t]
  \centering
  \includegraphics[width=1.0\textwidth]{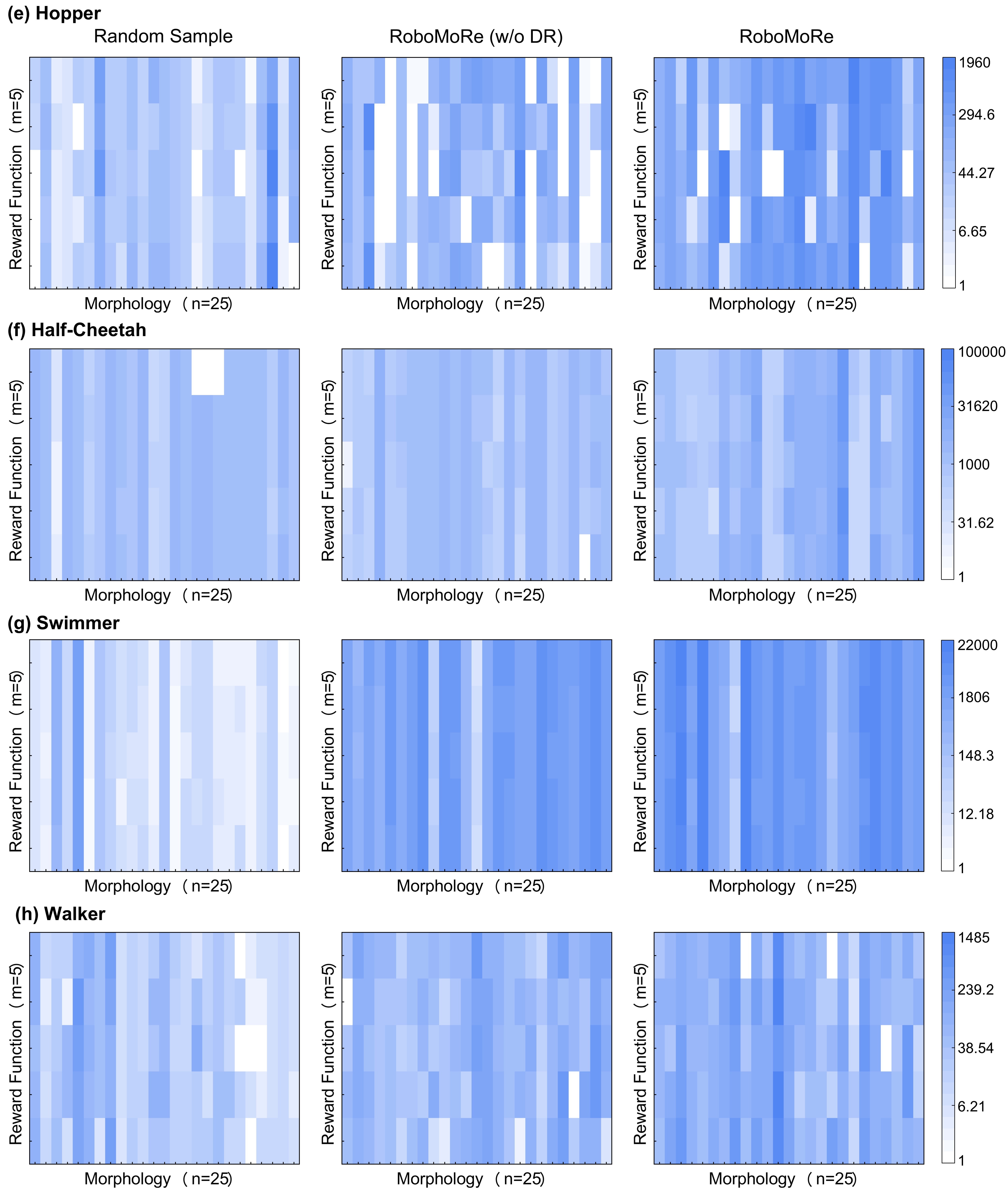} 
  \caption{\textbf{Coarse optimization results across 8 tasks (a–h) for Random Sample, RoboMoRe w/o Diversity Reflection (DR), and RoboMoRe.}}
  \label{Fig:CoarseA_H}
\end{figure*}

\subsection{Robustness across Different Gear Powers}
\label{supp:Robustness across Different Gear Powers}
\begin{figure*}[htbp]
  \centering
  \includegraphics[width=1.0\textwidth]{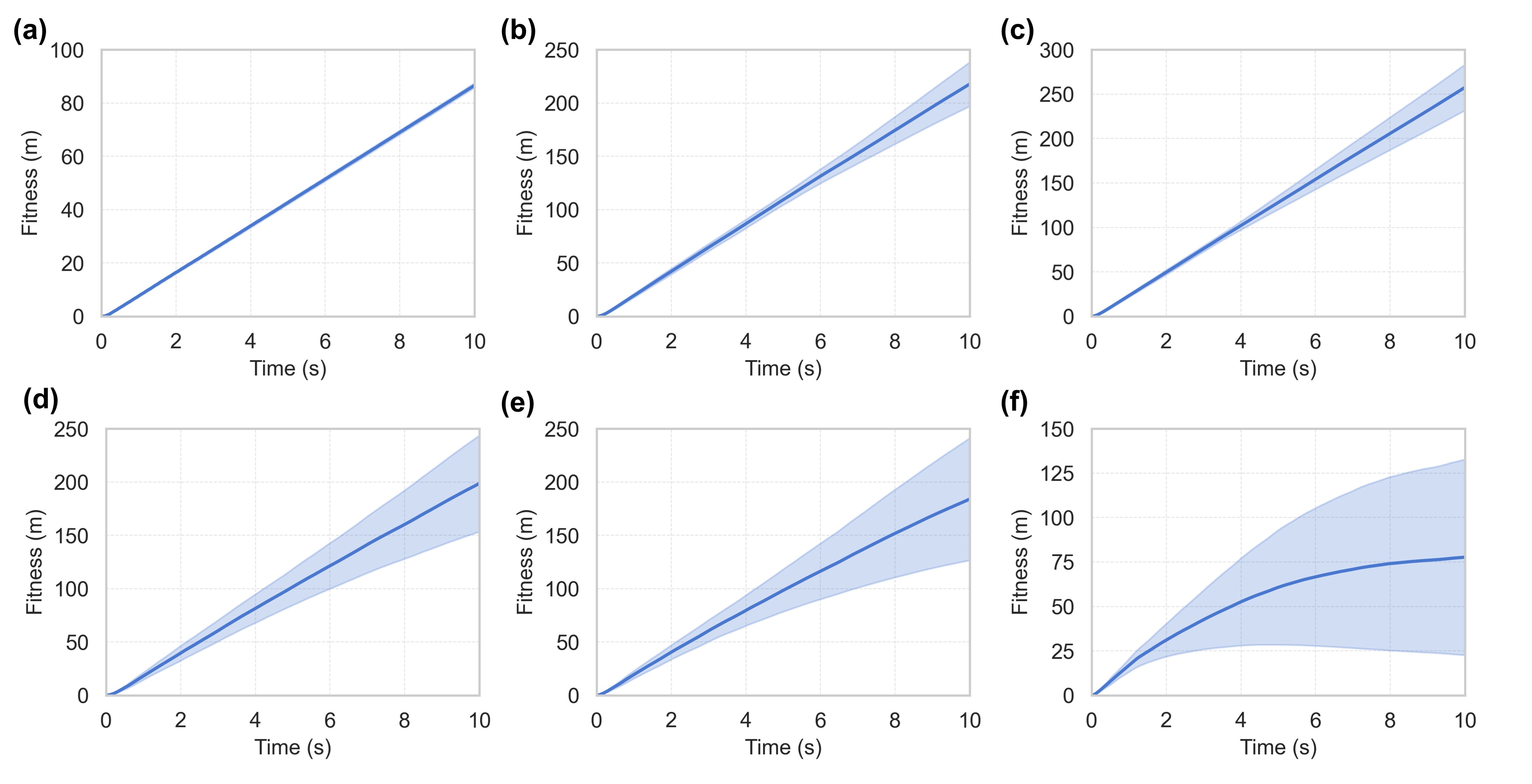} 
  \caption{ \textbf{Different power settings and corresponding results.} (a) Ant-Powered-50. (b) Ant-Powered-100. (c) Ant-Powered-150. (d) Ant-Powered-200. (e) Ant-Powered-250. (f) Ant-Powered-300.}
  \label{fig:Fig_power_experiment} 
\end{figure*}

To validate the robustness of RoboMoRe-generated designs across varying levels of motor power, we selected the top-performing design from the Ant-Powered task and retrained the model under five additional power settings. These power levels were configured directly through the MuJoCo XML files. The results, summarized in Fig.~\ref{fig:Fig_power_experiment}, show that the design discovered by RoboMoRe demonstrates strong adaptability and maintains consistent performance across a wide range of actuation capabilities.

\subsection{Robustness across Different Terrains}
\label{supp:Robustness across Different Terrains}
\begin{figure*}[htbp]
  \centering
  \includegraphics[width=1.0\textwidth]{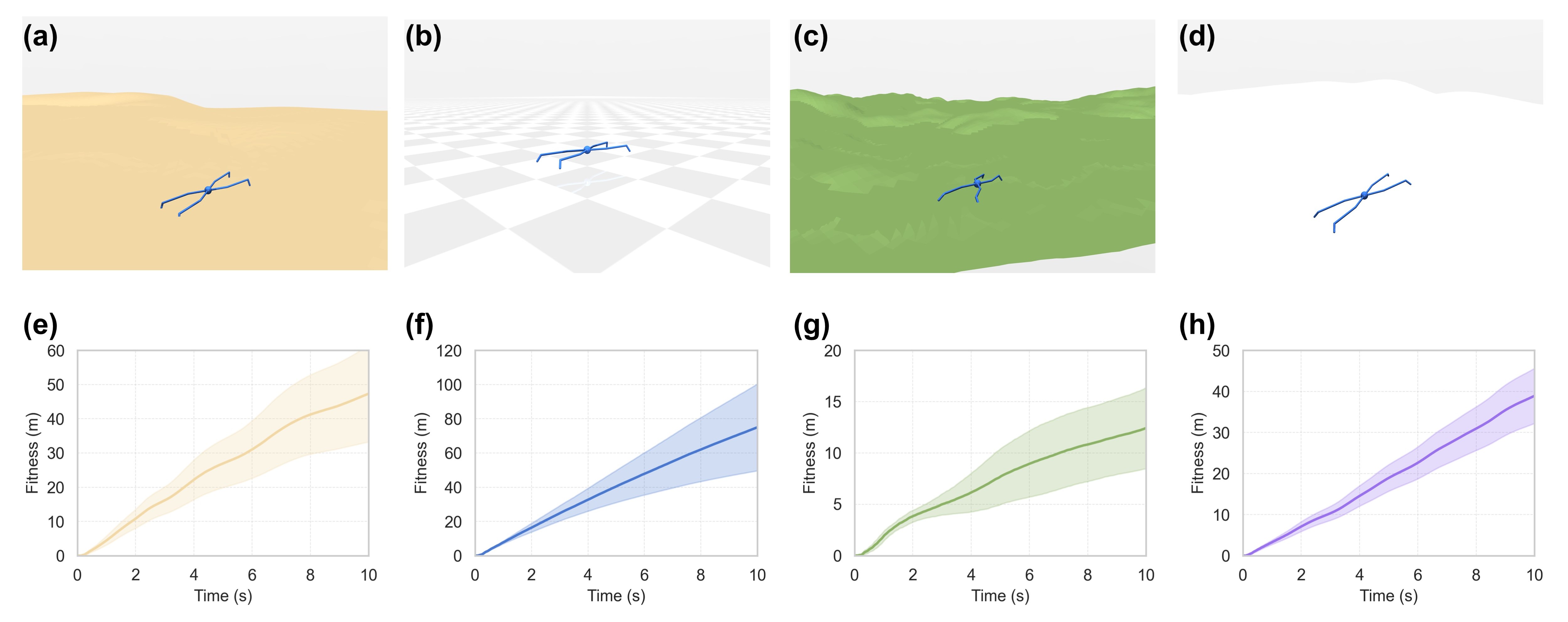} 
  \caption{\textbf{Different terrain environments and corresponding results.} (a) Ant-Desert. (b) Ant-Ground. (c) Ant-Hills. (d) Ant-Snow. (e) Results for Ant-Desert. (f) Results for Ant-Ground. (g) Results for Ant-Hills. (h) Results for Ant-Snow.}
  \label{fig:terrains} 
\end{figure*}

To further validate the robustness of RoboMoRe-generated designs across diverse terrains, we selected the top-performing design in Ant-Desert task and retrained the model on additional three distinct environments: ground, snow, and hills. These terrains were procedurally generated using Perlin noise and calibrated with realistic friction coefficients to reflect natural surface properties. The results, summarized in Fig. \ref{fig:terrains}, demonstrate that the design discovered by RoboMoRe exhibits strong adaptability and consistent performance across all tested terrains, highlighting its robustness in diverse physical conditions.

\clearpage
\newpage
\section{Algorithm Details}
\label{supp:algorithm}

\section{Algorithm Details}
\label{supp:algorithm}
\begin{algorithm}[htp]
\caption{\textbf{Coarse-to-Fine Optimization}}
\label{alg:robomore}
\KwData{
  Number of morphologies $\mathcal{N_M}$, number of rewards $\mathcal{N_R}$;\\
  Morphology prompt $P_{\mathcal{M}}$, reward prompt $P_{\mathcal{R}}$, diversity prompt $P_{\mathrm{diversity}}$; \\
  RL training algorithm $\mathrm{RL}(\cdot)$ (e.g., SAC); Fine optimization selection count $k$;
}
\KwResult{Optimized morphology $\theta^*$ and reward function $r^*$}
\BlankLine

\textcolor{gray}{\textit{// Coarse Optimization: Grid Search with Diversity Reflection.}}\;
$\Theta \gets \emptyset$\;
$\theta_1 \gets \mathrm{ProposeMorphology}(P_{\mathcal{M}})$\;
$\Theta \gets \Theta \cup \{\theta_1\}$\;

\For{$i \gets 2$ \KwTo $\mathcal{N_M}$}{
  $\theta_i \gets \mathrm{ProposeMorphology}(\Theta, P_{\mathcal{M}}, P_{\mathrm{diversity}})$\;
  $\Theta \gets \Theta \cup \{\theta_i\}$\;
}

$\mathcal{R} \gets \emptyset$\;
$r_1 \gets \mathrm{ProposeReward}(P_{\mathcal{R}})$\;
$\mathcal{R} \gets \mathcal{R} \cup \{r_1\}$\;

\For{$j \gets 2$ \KwTo $\mathcal{N_R}$}{
  $r_j \gets \mathrm{ProposeReward}(\mathcal{R}, P_{\mathcal{R}}, P_{\mathrm{diversity}})$\;
  $\mathcal{R} \gets \mathcal{R} \cup \{r_j\}$\;
}

$\mathcal{F} \gets \emptyset$\;
\ForEach{$\theta \in \Theta$}{
  \ForEach{$r \in \mathcal{R}$}{
    $F \gets \mathrm{RL}(\theta, r)$\;
    $\mathcal{F} \gets \mathcal{F} \cup \{(\theta, r, F)\}$\;
  }
}

$\mathcal{S}_{\text{coarse-best}} \gets \mathrm{Top}(\mathcal{F}, k)$\;

\BlankLine
\textcolor{gray}{\textit{// Fine Optimization: Alternating Morphology and Reward Optimization}}\;
$\mathcal{S}_{\text{fine}} \gets \emptyset$\;

\ForEach{$(\theta, r) \in \mathcal{S}_{\text{coarse-best}}$}{
  $\theta^* \gets \theta$, $r^* \gets r$\;
  $F^* \gets \mathrm{RL}(\theta^*, r^*)$\;
  \While{not converged}{
    improved $\gets$ False\;

    \textcolor{gray}{\textit{// Morphology Optimization}}\;
    $\theta' \gets \mathrm{ImproveMorphology}(\theta^*,r^*,\Theta,\mathcal{F})$\;
    $F' \gets \mathrm{RL}(\theta', r^*)$\;
    \If{$F' > F^*$}{
      $\theta^* \gets \theta'$, $F^* \gets F'$;\;
      improved $\gets$ True;\;
    }

    \textcolor{gray}{\textit{// Reward Function Optimization}}\;
    $r' \gets \mathrm{ImproveReward}(\theta^*,r^*,\mathcal{R}, \mathcal{F})$\;
    $F' \gets \mathrm{RL}(\theta^*, r')$\;
    \If{$F' > F^*$}{
      $r^* \gets r'$, $F^* \gets F'$;\;
      improved $\gets$ True;\;
    }

    \If{not improved}{
      break\;
    }
  }

  $\mathcal{S}_{\text{fine}} \gets \mathcal{S}_{\text{fine}} \cup \{(\theta^*, r^*, F^*)\}$\;
}
\Return $\text{Best } (\theta^*, r^*) \text{ from } \mathcal{S}_{\text{fine}}$\;

\end{algorithm}

\newpage
\section{More Visualization Results}
\label{supp:Comparison of optimal morphology design via different methods}
\subsection{Comparison of optimal morphology design via different methods}
\begin{figure*}[htbp]
  \centering
  \includegraphics[width=1.0\textwidth]{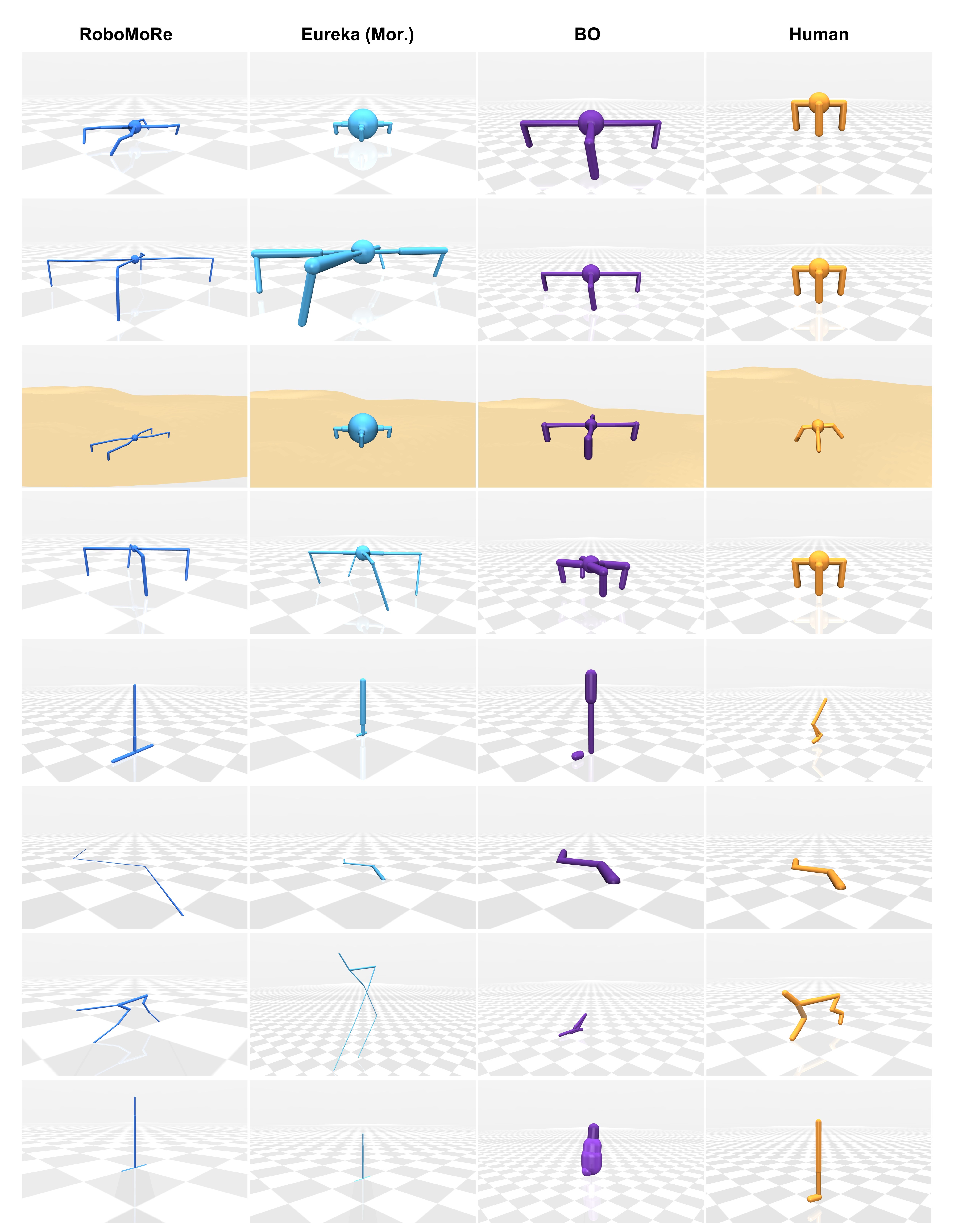} 
  \caption{\textbf{Optimized morphology with comparison methods.} }
  \label{fig:optimized_morphology} 
\end{figure*}

Fig.~\ref{fig:optimized_morphology} presents the optimized morphology design. Evidently, RoboMoRe is capable of producing highly efficient structures, demonstrating its effectiveness in morphology optimization.

\clearpage
\subsection{Ant: Best Reward Function}
\begin{lstlisting}[style=pythoncolorstyle,label={1st:task_description}]
import numpy as np
def _get_rew(self, x_velocity: float, action):
    # Encourage not just moving forward but also involving some controlled lateral movement 
    # to demonstrate agility and dynamic control capability.
    
    # Reward for moving forward
    forward_reward = self._forward_reward_weight * x_velocity
    
    # Reward for controlled lateral movement (moderate y_velocity to demonstrate lateral agility)
    target_y_velocity = 0.5  # moderate lateral speed target, can be tuned
    lateral_movement_reward = -np.abs(self.data.qvel[1] - target_y_velocity) * self._forward_reward_weight
    
    # Minimize the control effort to promote energy efficiency.
    control_cost = self.control_cost(action)
    
    # Health reward for maintaining a physically feasible and stable posture.
    health_reward = self.healthy_reward
    
    # Total reward computation
    reward = forward_reward + lateral_movement_reward - control_cost + health_reward
    
    # Reward details for debugging and analysis purposes
    reward_info = {
        'forward_reward': forward_reward,
        'lateral_movement_reward': lateral_movement_reward,
        'control_cost': control_cost,
        'health_reward': health_reward
    }

    return reward, reward_info
\end{lstlisting}

\subsection{Ant-Powered: Best Reward Function}
\begin{lstlisting}[style=pythoncolorstyle,label={1st:task_description}]
import numpy as np 
def _get_rew(self, x_velocity: float, action):
    # Encourage forward movement with a target velocity for optimal speed
    target_velocity = 1.0  # Desired forward velocity
    velocity_error = np.abs(x_velocity - target_velocity)  # Calculate deviation from target
    forward_reward = self._forward_reward_weight * np.exp(-velocity_error)  # Exponential decay for reward

    # Penalize lateral (y-direction) movement to maintain focus on forward motion
    y_velocity_penalty = -abs(self.data.qvel[1])  # Penalize any motion in the y direction

    # Introduce a reward for efficient alternate leg movement to promote stability and locomotion style
    leg_pair_1 = np.abs(action[0] + action[1] - (action[4] + action[5]))
    leg_pair_2 = np.abs(action[2] + action[3] - (action[6] + action[7]))
    alternate_leg_movement_reward = (np.exp(-leg_pair_1) + np.exp(-leg_pair_2)) / 2.0

    # Maintain healthy posture by rewarding stability within the Z-limits
    health_reward = self.healthy_reward

    # Evaluate the efficiency of movement through control costs
    control_cost = self.control_cost(action)
    
    # Penalize excessive contact forces to minimize instability 
    contact_cost = self.contact_cost

    # Calculate total reward by combining all components
    reward = forward_reward + y_velocity_penalty + alternate_leg_movement_reward - control_cost + health_reward - contact_cost

    # Reward info for monitoring individual components
    reward_info = {
        "forward_reward": forward_reward,
        "y_velocity_penalty": y_velocity_penalty,
        "alternate_leg_movement_reward": alternate_leg_movement_reward,
        "control_cost": control_cost,
        "contact_cost": contact_cost,
        "health_reward": health_reward,
    }

    return reward, reward_info
\end{lstlisting}
\clearpage
\subsection{Ant-Desert: Best Reward Function}
\begin{lstlisting}[style=pythoncolorstyle,label={1st:task_description}]
import numpy as np 
def _get_rew(self, x_velocity: float, action):
    # Reward for forward movement, emphasizing an exponential growth to encourage high speeds.
    forward_reward = self._forward_reward_weight * np.exp(x_velocity - 1)  # Exponential growth for incentivizing high speeds

    # Encourage lateral stability: penalize the absolute value of lateral (y-axis) velocity
    y_velocity = self.data.qvel[1]
    lateral_stability_penalty = -np.abs(y_velocity)  # Negative as we want to minimize lateral movement

    # Control cost to penalize excessive energy usage
    control_cost = self.control_cost(action)

    # Keep the robot in a healthy state
    healthy_reward = self.healthy_reward

    # Minimize contact costs to promote gentle contacts with the ground
    contact_cost = self.contact_cost

    # Total reward calculation combining all components
    reward = forward_reward + healthy_reward + lateral_stability_penalty - control_cost - contact_cost

    # Detailed reward breakdown for analysis and debugging
    reward_info = {
        'forward_reward': forward_reward,
        'lateral_stability_penalty': lateral_stability_penalty,
        'control_cost': control_cost,
        'contact_cost': contact_cost,
        'healthy_reward': healthy_reward
    }

    return reward, reward_info
\end{lstlisting}

\subsection{Ant-Jump: Best Reward Function} 
\begin{lstlisting}[style=pythoncolorstyle,label={1st:task_description}]
import numpy as np 
def _get_rew(self, x_velocity: float, action):
    # Decomposing the reward function
    jump_height = self.data.body(self._main_body).xpos[2]  # Z position gives the height
    
    # Reward for jumping higher
    height_reward = np.exp(jump_height - 1)  # exponential reward starting from height 1
    
    # Control cost to make sure the robot uses minimum effort to jump
    control_cost = self.control_cost(action)
    
    # Contact cost to penalize excessive force usage in contacts, promoting smooth jumping
    contact_cost = self.contact_cost
    
    # Component to support healthy posture
    healthy_posture_reward = self.healthy_reward
    
    # Combination of different components of the reward
    reward = (
        height_reward * self._forward_reward_weight 
        - contact_cost 
        - control_cost 
        + healthy_posture_reward
    )
    
    # Reward info for better analysis and debugging
    reward_info = {
        "height_reward": height_reward,
        "control_cost": control_cost,
        "contact_cost": contact_cost,
        "healthy_posture_reward": healthy_posture_reward,
    }
    
    return reward, reward_info
\end{lstlisting}
\clearpage
\subsection{Hopper: Best Reward Function}
\begin{lstlisting}[style=pythoncolorstyle,label={1st:task_description}]
import numpy as np
def _get_rew(self, x_velocity: float, action):
    # Encourage efficient forward movement by not just rewarding speed but also smooth progression
    smoothness_reward = -np.sum(np.abs(np.diff(action)))  # Decrease reward for large differences in consecutive actions

    # Reward forward velocity, with an exponential component to prioritize higher speeds exponentially
    exponential_speed_reward = self._forward_reward_weight * np.exp(x_velocity) - 1  # Using exp to exponentially prefer higher velocities, subtract 1 to center around zero for small velocities

    # Penalty for using too much control input, which promotes efficiency
    control_penalty = self._ctrl_cost_weight * np.sum(np.square(action))

    # Healthy state reward, keeping the hopper upright and in a healthy range
    health_bonus = self.healthy_reward

    # Total reward calculation
    total_reward = exponential_speed_reward + smoothness_reward - control_penalty + health_bonus

    # Tracking reward details for better understanding and debugging
    reward_info = {
        'smoothness_reward': smoothness_reward,
        'exponential_speed_reward': exponential_speed_reward,
        'control_penalty': control_penalty,
        'health_bonus': health_bonus,
        'total_reward': total_reward
    }

    return total_reward, reward_info
\end{lstlisting}

\subsection{Half-Cheetah: Best Reward Function}
\begin{lstlisting}[style=pythoncolorstyle,label={1st:task_description}]
import numpy as np 
def _get_rew(self, x_velocity: float, action):
    # Reward for moving forward emphasizing higher speeds
    forward_reward = self._forward_reward_weight * x_velocity

    # Calculate control cost using the predefined method
    control_cost = self.control_cost(action)

    # Reward for energy efficiency: velocity per control effort
    efficiency = x_velocity / (control_cost + 1e-5)  # Avoid division by zero
    normalized_efficiency_reward = np.exp(efficiency) - 1  # Shifted by -1 to normalize around 0

    # Calculate smoothness reward: penalize fluctuations in velocity
    if not hasattr(self, 'prev_velocity'):
        self.prev_velocity = x_velocity
    smoothness_penalty = -np.abs(x_velocity - self.prev_velocity)  # Penalize changes in velocity
    self.prev_velocity = x_velocity  # Update the previous velocity for the next step
    smoothness_reward = np.exp(smoothness_penalty) - 1  # Normalize the smoothness reward
    
    # Action symmetry bonus: rewards symmetrical actions between limbs
    if len(action) % 2 == 0:  
        left_actions = action[1::2]  
        right_actions = action[0::2]  
        symmetry_penalty = -np.sum(np.abs(left_actions - right_actions))  
    else:
        symmetry_penalty = 0  
    symmetry_reward = np.exp(symmetry_penalty) - 1  
    
    # Combine all components to form the total reward
    total_reward = forward_reward - control_cost + normalized_efficiency_reward + smoothness_reward + symmetry_reward

    # Reward info dictionary for debugging and analysis
    reward_info = {
        'forward_reward': forward_reward,
        'control_cost': control_cost,
        'normalized_efficiency_reward': normalized_efficiency_reward,
        'smoothness_reward': smoothness_reward,
        'symmetry_reward': symmetry_reward,
        'total_reward': total_reward
    }

    return total_reward, reward_info
\end{lstlisting}

\subsection{Swimmer: Best Reward Function}
\begin{lstlisting}[style=pythoncolorstyle,label={1st:task_description}]
import numpy as np 
def _get_rew(self, x_velocity: float, action):
    # Define efficient swimming characteristics. Aim to maintain balance between forward motion and control.
    
    # Base forward reward for moving right, placed higher for higher velocities
    forward_reward = self._forward_reward_weight * (x_velocity ** 2)  # Quadratic reward amplifies strong velocities

    # Introduce a reward for stability: Encourage minimal yawing (side ways movements).
    stability_reward = -0.5 * (self.data.qpos[1] ** 2)  # Penalizing lateral position squared to discourage side motion

    # Punishing excessive control inputs, to favor smooth swimming rather than jerky movements
    control_penalty = self.control_cost(action)

    # Incentivize maintaining a certain baseline velocity (e.g., cruising speed), with a soft penalty for deviations
    ideal_velocity = 1.0
    velocity_penalty = 0.5 * ((x_velocity - ideal_velocity) ** 2)  # Penalizes big deviations from ideal

    # Compute the total reward
    total_reward = forward_reward + stability_reward - control_penalty - velocity_penalty

    # Debugging and analysis information containing detailed components of the reward
    reward_info = {
        'forward_reward': forward_reward,
        'stability_reward': stability_reward,
        'control_penalty': control_penalty,
        'velocity_penalty': velocity_penalty,
    }
    
    return total_reward, reward_info
\end{lstlisting}

\subsection{Walker: Best Reward Function}
\begin{lstlisting}[style=pythoncolorstyle,label={1st:task_description}]
import numpy as np
def _get_rew(self, x_velocity: float, action):
    # Reward for forward motion, scaled exponentially to favor higher speeds but with diminishing returns
    forward_reward = np.exp(0.3 * x_velocity) - 1

    # Penalty to discourage excessive use of actuator torques, promoting energy efficiency
    control_penalty = self.control_cost(action)
    
    # Bonus for maintaining a healthy mechanics of motion, which includes staying upright    
    health_bonus = self.healthy_reward
    
    # Additional reward for synchronous and rhythmic leg movements
    # We can leverage the sine of the sum of relevant joint angles to promote a smooth cyclic locomotion
    angles = self.data.qpos[2:7]  # Assuming indices 2-6 are joint angles
    rhythmic_movement_bonus = np.sum(np.sin(angles))

    # Compute total reward considering all the components
    reward = forward_reward + rhythmic_movement_bonus + health_bonus - control_penalty
    
    # Organize detailed reward components for diagnostic purposes
    reward_info = {
        'forward_reward': forward_reward,
        'rhythmic_movement_bonus': rhythmic_movement_bonus,
        'health_bonus': health_bonus,
        'control_penalty': control_penalty
    }
    
    return reward, reward_info
\end{lstlisting}

\end{document}